\documentclass[a4paper,conference]{IEEEtran}
\pdfoutput=1
\usepackage[hyperref,dvipsnames]{xcolor}
\usepackage[pdftex]{graphicx}
\graphicspath{{\rootpath/img/}}
\usepackage{subfig}

\usepackage{todonotes}

\usepackage{xparse}
\usepackage{grffile}

\usepackage{amsmath}
\usepackage{amssymb}
\usepackage{array}
\usepackage{url}

\usepackage{hyperref}
\newcommand\myshade{75}
\hypersetup{
    colorlinks=true,
    linkcolor=violet!\myshade!black,
    citecolor=YellowOrange!\myshade!black,
    urlcolor=Aquamarine!\myshade!black,
}

\DeclareMathOperator*{\argmax}{arg\,max}
\DeclareMathOperator*{\argmin}{arg\,min}

\newcommand{\w}{\mathbf{w}}

\NewDocumentCommand {\Prob}{ m o } {%
  \IfValueTF{#2}{%
    \mathit{P} ( #1 | #2  )%
    }{%
    \mathit{P}  ( #1  )%
    }%
}
\newcommand{\D}{\mathcal{D}}
\NewDocumentCommand {\Prior}{o} {\Prob{\w \IfValueT{#1}{^{(#1)}}}}
\NewDocumentCommand {\Post}{o} {\Prob{\w \IfValueT{#1}{^{(#1)}}}[\D]}
\NewDocumentCommand {\Lhod}{o} {\Prob{\D}[\w \IfValueT{#1}{^{(#1)}}]}
\NewDocumentCommand {\pt}{} {\mathbf{p}_t}
\newcommand{\xhat}{\hat{\mathbf{x}}}
\newcommand{\yhat}{\hat{\mathbf{y}}}
\NewDocumentCommand{\q}{o}{\mathit{q} ( \w \IfValueT{#1}{^{(#1)}} | \theta  )}
\newcommand{\intdw}[1]{\int #1 \, d \w}

\newcommand{\tsdfwidth}{1.6in}
\newcommand{\rootpath}{.}

\begin{document}

\title{In Depth Bayesian Semantic Scene Completion}
% \title{Bayesian Semantic Scene Completion from a single depth image}

\author{\IEEEauthorblockN{David Gillsj\"o}
\IEEEauthorblockA{Centre for Mathematical Sciences\\
Lund University\\
david.gillsjo@math.lth.se}
\and
\IEEEauthorblockN{Kalle \AA str\"om}
\IEEEauthorblockA{Centre for Mathematical Sciences\\
Lund University\\
karl.astrom@math.lth.se}
}

%Telephone: (800) 555--1212, Fax: (888) 555--1212}

\maketitle
\begin{abstract}
For autonomous agents moving around in our world, mapping of the environment is essential.
This is their only perception of their surrounding, what is not measured is unknown.
Humans have learned from experience what to expect in certain environments,
for example in indoor offices or supermarkets.

This work studies Semantic Scene Completion which aims to predict a 3D semantic segmentation
of our surroundings, even though some areas are occluded.
For this we construct a Bayesian Convolutional Neural Network (BCNN), which is
not only able to perform the segmentation, but also predict model uncertainty.
This is an important feature not present in standard CNNs.

We show on the MNIST dataset that the Bayesian approach performs equal or better
to the standard CNN when processing digits unseen in the training phase when looking at accuracy, precision and recall.
With the added benefit of having better calibrated scores and the ability to express model uncertainty.

We then show results for the Semantic Scene Completion task
where a category is introduced at test time on the SUNCG dataset.
In this more complex task the Bayesian approach outperforms the standard CNN.
Showing better Intersection over Union score and excels in Average Precision and separation scores.

\end{abstract}
\section{Introduction}
Semantic scene completion is a challenging task in which both visible and occluded surfaces are labeled semantically in 3D.
The problem naturally arises when aiming to predict a 3D scene from a single view, but can also be studied for multiple views.

\begin{figure}[!t]
\centering
\includegraphics[width=.45\textwidth]{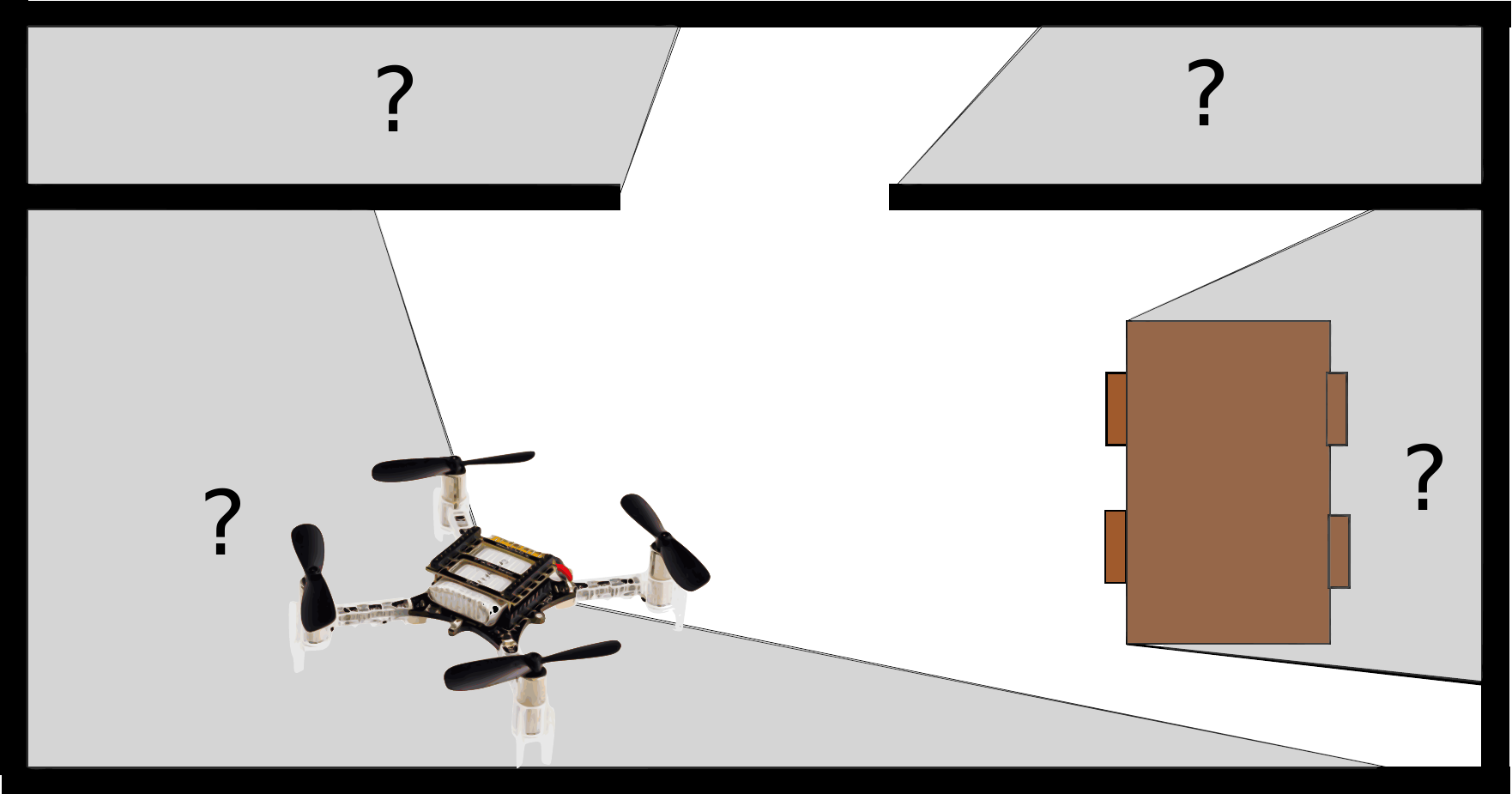}%
\caption{An UAV has some occluded areas in its surrounding and would like to have an idea about what to expect.}
\label{fig:motivation}
\end{figure}

Predicting occluded areas can be of great help to autonomous vehicles during navigation and exploration.
Especially for UAVs (Unmanned Aerial Vehicles), which navigate a 3D space where observations may be sparse, scene completion can be used for smoother trajectories during path planning.
During exploration it can help the agent understand the likelihood of free space in occluded areas.
In Figure \ref{fig:motivation} we see an illustration of the problem where a UAV would benefit from knowing what to expect in occluded areas.

This work introduces a system for Bayesian Semantic Scene Completion (BSSC), which along with the prediction scores also delivers an estimation of uncertainty.
This is crucial for decision making during autonomous navigation and exploration as it can
help the agent understand when the data is new to the model and the prediction should not be trusted.
It can also be used to understand what data should be added to the training to improve robustness.

We first test our implementation on the MNIST dataset to verify the Bayesian approach and
to understand how the output distributions look for a well modeled dataset.
We then move on to the SUNCG data for the Semantic Scene Completion task.

Our contributions include:
\begin{itemize}
  \item An open source system for BSSC using Variational Inference released on \url{https://github.com/DavidGillsjo/bssc-net}.
  \item An extended SSC task on the SUNCG dataset which includes more occluded space.
  \item Experiments showing that the Bayesian approach is more robust to unseen data in the SSC task.
  \item Parameter studies on both MNIST and SUNCG.
\end{itemize}

\section{Related work} \label{sec:related}
Semantic scene completion has been formulated for both the single view and multiple view problem with different sensor modalities.
One example is SSC-Net \cite{song2016ssc} that solves the single view using a 3D CNN with depth data as input. Our work is heavily inspired by their architecture and training setup.
Other works \cite{Garbade_2019_CVPR_Workshops,Liu_NIPS2018} have then extended the architecture by utilizing the RGB information as well.

Since the solutions are similar to Semantic Scene Segmentation, the following works are also of interest.
Rather than feeding the whole input volume through the network, \cite{Dai_2018_CVPR} takes a sliding window approach to better handle large datasets.
By encoding the RGB information as features and using differentiable backprojection 3DMV \cite{dai20183dmv} got impressive results on the ScanNet dataset \cite{dai2017scannet}.

The best performing networks on ScanNet semantic segmentation is currently based on point clouds \cite{SparseConvNet_CVPR2018,thomas2019KPConv,wu2018pointconv} or sparse convolutions \cite{Choy_CVPR2019}.

Our work will not focus on performance, but rather how to estimate uncertainty in semantic scene completion using Bayesian NN.

There are several ways of modeling a Bayesian Neural Network, this work uses the Bayes by Backpropagation \cite{blundell2015weight}.
Our implementation of Bayes by Backpropagation is based on \cite{shridhar2019comprehensive} which implements this for a 2D CNN using softplus and normalized softplus as activation functions.

Both occupancy maps and semantic segmentation typically use a probabilistic representation where each class is assigned probability between 0 and 1.
If the algorithm also outputs uncertainty together with the probability score, what does that mean and how do we use it?
This is studied for Gaussian Process Occupancy Maps (GPOM) \cite{gpom} in the robotics community.
GPOM yields a mean and variance per grid cell, these are then fed to a linear classifier, which is trained to output a probability score.

% By dividing the uncertainty estimate into Aleatoric and Epistemic uncertainty \cite{kendall2017uncertainties},
% which capture measurement noise and model noise respectively,
% it is possible to understand what can be improved by adding new data to the training process.

Most NN classifiers does not have calibrated probabilities, this can be adjusted with for example Platt Scaling as in \cite{kuleshov2018accurate}.
Uncertainties from Bayesian NN classifier can be similarly calibrated.
In \cite{seedat2019calibrated} the authors study calibration for different BNNs by comparing Entropy, Mutual Information, Aleatoric uncertainty and Epistemic uncertainty
under different perturbations of input data. They conclude that predictive entropy and Epistemic uncertainty provide the most robust uncertainty estimates.
%They also get poor result using VI compared to the other methods.

%
\section{Bayes by backprop}
\begin{figure}[!t]
\centering
\subfloat[Standard]{\includegraphics[width=.46\linewidth, trim=300 150 100 150, clip]{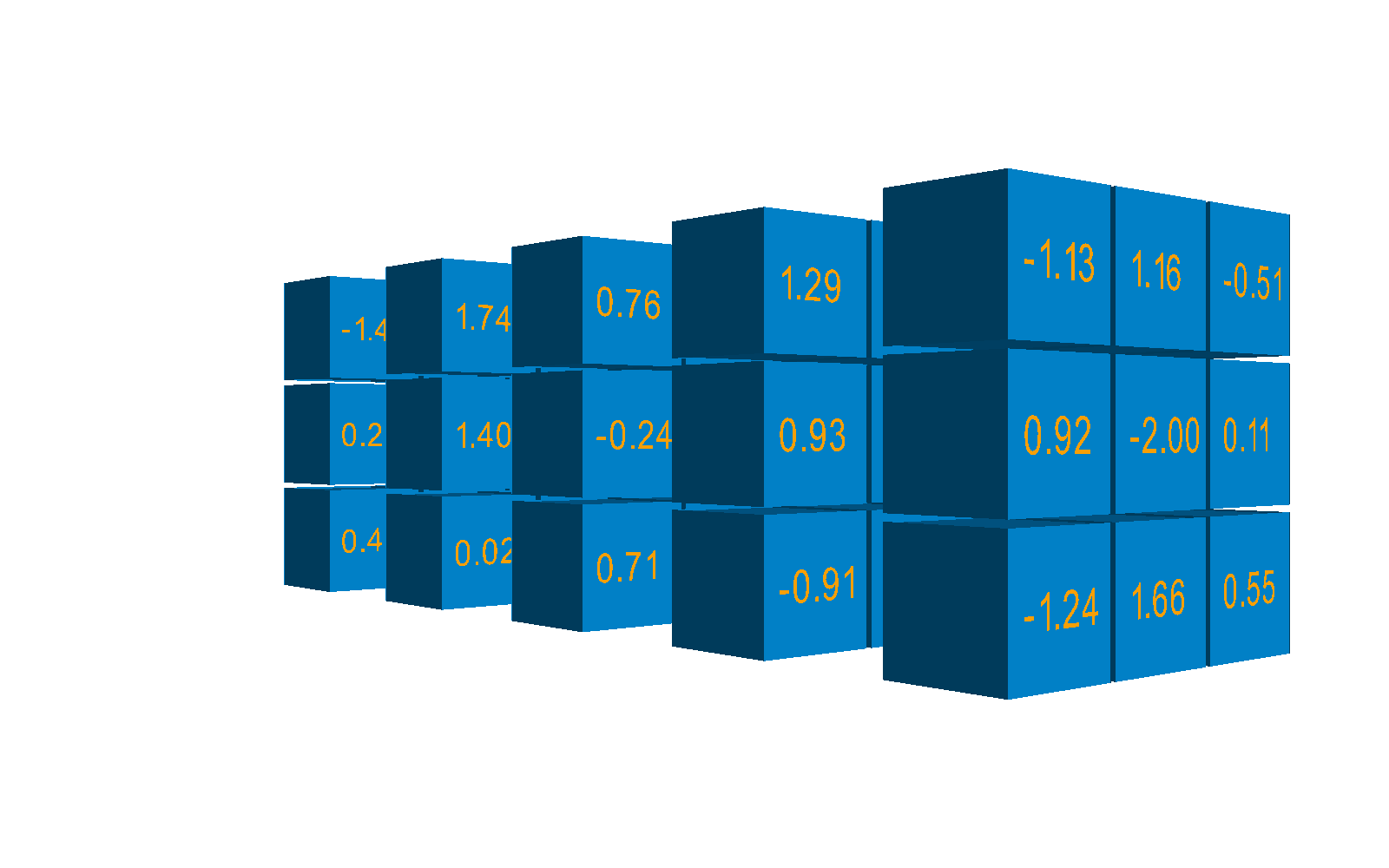}%
\label{fig:standard}}
\hfil
\subfloat[Bayesian]{\includegraphics[width=.46\linewidth, trim=300 150 100 150, clip]{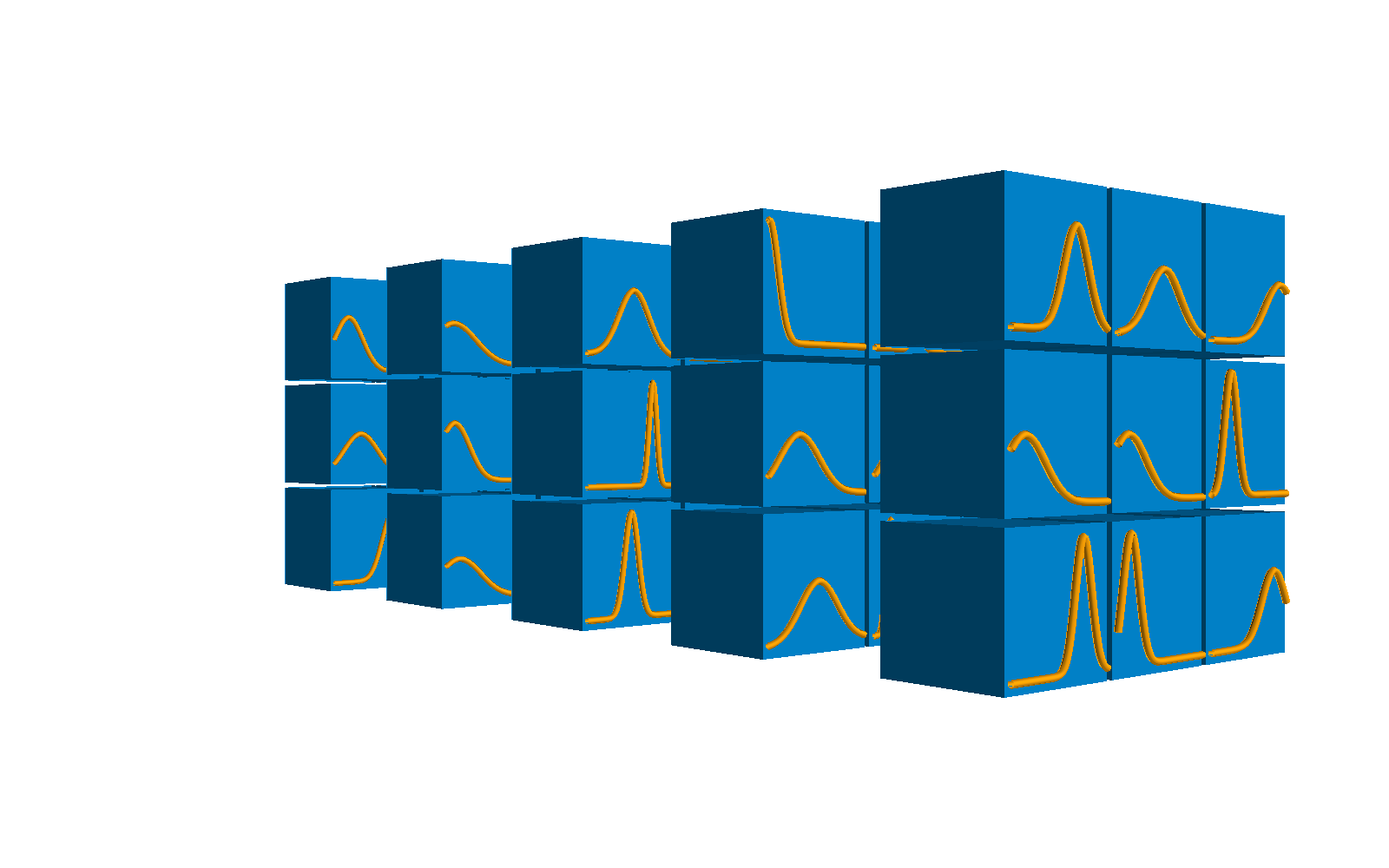}%
\label{fig:bayesian}}
\caption{In \ref{fig:standard} we see a filter bank from a standard 2D CNN, each weight is a scalar.
In \ref{fig:bayesian} we see a filter bank in a Bayesian Variational Inference 2D CNN, here each weight represented as a distribution which is sampled from at inference time.}
\label{fig:bayesian_vs_standard}
\end{figure}

The method introduced by \cite{blundell2015weight} is based on Variational Inference.
The main idea is to let each weight in the network be sampled from a distribution,
where the distribution is learned at training time, as illustrated in Figure \ref{fig:bayesian_vs_standard}.

Let $\Post$ be the posterior for our Bayesian Neural Network.
Given the posterior we can make predictions given unseen data by taking the expectation over the posterior
\begin{align*}
    \Prob{\yhat}[\xhat] &= \mathbb{E}_{\Post} \left [ \Prob{\yhat}[\xhat, \w] \right ]\\
                        &= \intdw{ \Prob{\yhat}[\xhat, \w] \Post }.
\end{align*}
We estimate the posterior using a simpler model $\q$ with learnable parameters $\theta$,
which minimizes the Kullback-Leibler (KL) divergence to the true posterior, i.e.
\begin{align*}
  \theta^* &= \argmin_\theta KL[\q || \Post] \\
           &= \argmin_\theta \int \q \log \frac{\q}{\Prior \Lhod} \\
           &= \argmin_\theta KL[\q || \Prior] -  \mathbb{E}_{\q} [\log \Lhod].
\end{align*}
This cost function is known as variational free energy or expected lower bound.
We denote it as
\begin{equation} \label{eq:F_cost}
  \mathcal{F}(\D,\theta) = KL[\q || \Prior] -  \mathbb{E}_{\q} [\log \Lhod].
\end{equation}
Where the first part is the complexity cost as it enforces model simplicity using the prior.
The second part is the likelihood, which describes how well the model describes the data.

Using Monte Carlo sampling the cost in (\ref{eq:F_cost}) is approximated as
\begin{align*} \label{eq:F_cost_approx}
  \mathcal{F}(\D,\theta) \approx &\sum_{i=1}^n \frac{\beta}{n} \left [ \log \q[i] - \log \Prior[i] \right ]\\
                                 &- \log \Lhod[i],
\end{align*}
where $\w^{(i)}$ is a sample from the variational posterior $\q[i]$.
This approximation enables more priors and posteriors since a closed for solution for the first term in (\ref{eq:F_cost}) is not necessary.
The scale factor $\frac{\beta}{n}$ with $\beta$ as design parameter is introduced to tune the amount of regularization from the complexity cost.
\subsection{Variational Posterior}
Assuming the variational posterior $\q$ is a diagonal Gaussian distribution.
With the re-parametrization trick \cite{reparametrization} the weights can be sampled from the posterior as
\begin{equation*}
  \w = \mu + \log (1 + e^\gamma) \odot \epsilon, \quad \epsilon \sim \mathcal{N}(0,I),
\end{equation*}
where the model parameters are $\theta = (\mu, \gamma)$ and $\sigma = \log (1 + e^\gamma)$
to ensure that the standard deviation remains positive during optimization.
\subsection{Prior}
For the choice of prior we tested both Gaussian (as in \cite{shridhar2019comprehensive}), Cauchy and a scale mixture of two Gaussian (as in \cite{blundell2015weight}).
The two latter affords more flexibility to the network and we found they yielded the best results.
We mainly worked with Cauchy since it is more efficient to compute but have some results on the Gaussian scale mixture.
As they are centered around 0 we denote them as
\begin{equation*}
  \text{Cauchy}(\gamma) = \frac{1}{\pi \gamma \left [1 + (\frac{x}{\gamma})^2 \right]}
\end{equation*}
and
\begin{equation*}
  \text{G}^M_{(\sigma_0, \sigma_1)}(\alpha) = \alpha \mathcal{N}(0,\sigma_0^2) + (1-\alpha) \mathcal{N}(0,\sigma_1^2),
\end{equation*}
where $\mathcal{N}$ denotes the probability density function of the normal distribution.
\subsection{Prediction \& Uncertainty}
\label{sec:prediction}
The predictive mean and uncertainty is computed by sampling our variational posterior (BCNN) at test time.
An unbiased estimation of the expectation is given \cite{shridhar2019comprehensive} by
\begin{align*}
  \mathbb{E}_{\q} \left [ \Prob{\yhat}[\xhat, \w] \right ] &=  \intdw{\q \Prob{\yhat}[\xhat,\w^{(t)}]} \\
                                                           & \approx \frac{1}{T} \sum^T_{t=1} \Prob{\yhat}[\xhat,\w^{(t)}],
\end{align*}
where $\Prob{\yhat}[\xhat,\w^{(t)}]$ is the softmax output from forward pass $t$.
To simplify notation we now denote this as $\pt$.

There are a number of choices when it comes to measuring the uncertainty \cite{seedat2019calibrated,kendall2017uncertainties}.
Common choices are predictive entropy, aleatoric uncertainty and epistemic uncertainty.

\textbf{Predictive Entropy -} Measures the diversity of the distribution, so a higher
entropy corresponds to higher uncertainty. It is computed as
\begin{equation*}
  H = - \sum^T_{t=1} \pt \log \pt.
\end{equation*}

\textbf{Aleatoric Uncertainty -} Corresponds to measurement noise from the input data,
so increasing the size of the dataset should not reduce this uncertainty. It is computed as
\begin{equation*}
  \sigma^2_a = \frac{1}{T} \sum^T_{t=1} \text{diag} ( \pt ) - \pt \pt^\top.
\end{equation*}

\textbf{Epistemic Uncertainty -} Corresponds to model uncertainty and will be low when
input data is similar to training data. Increasing the amount and diversity of training data should reduce this uncertainty.
This is computed as
\begin{equation*}
  \sigma^2_e = \frac{1}{T} \sum^T_{t=1} ( \pt - \bar{\pt} )( \pt - \bar{\pt} )^\top,
\end{equation*}
where $\bar{\pt} = \frac{1}{T}  \sum^T_{t=1} \pt $ is the predictive mean from above.

\section{Network architecture}
\begin{figure*}[!t]
\centering
\subfloat[MNIST]{\includegraphics[width=.22\textwidth]{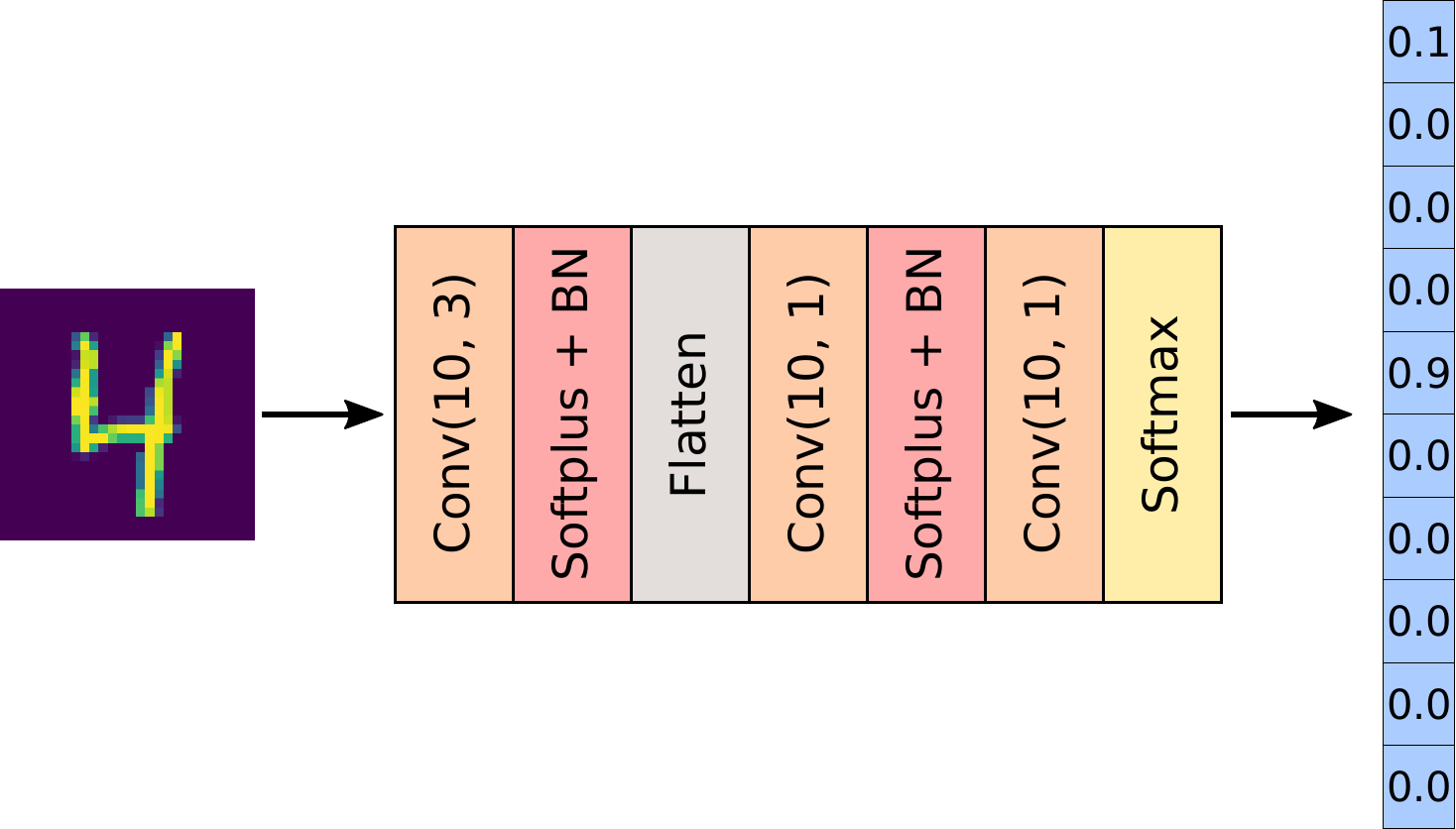}%
\label{fig:MNIST_arch}}
\hfil
\subfloat[SSC-Net]{\includegraphics[width=.6\textwidth]{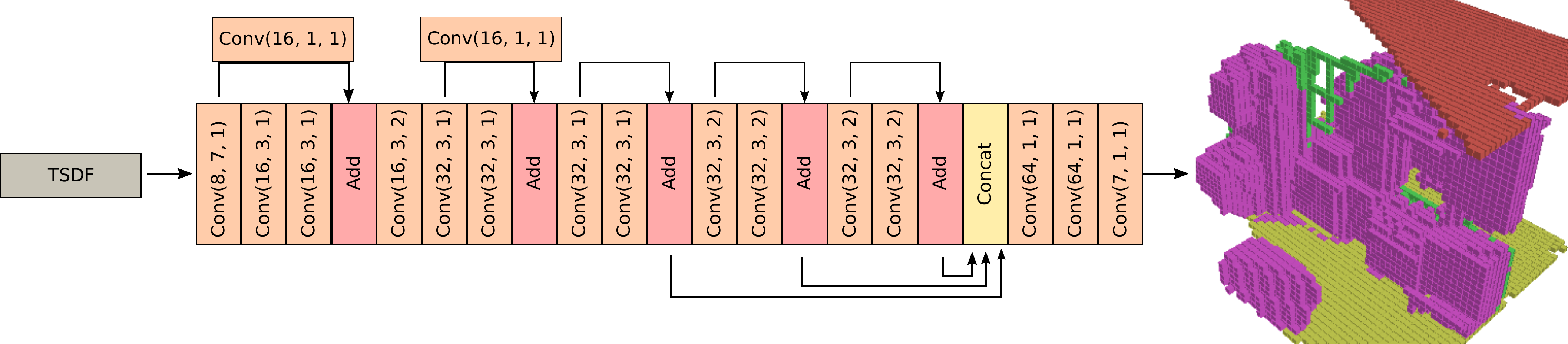}%
\label{fig:SUNCG_arch}}
\hfil
\subfloat[UNet]{\includegraphics[width=.7\textwidth]{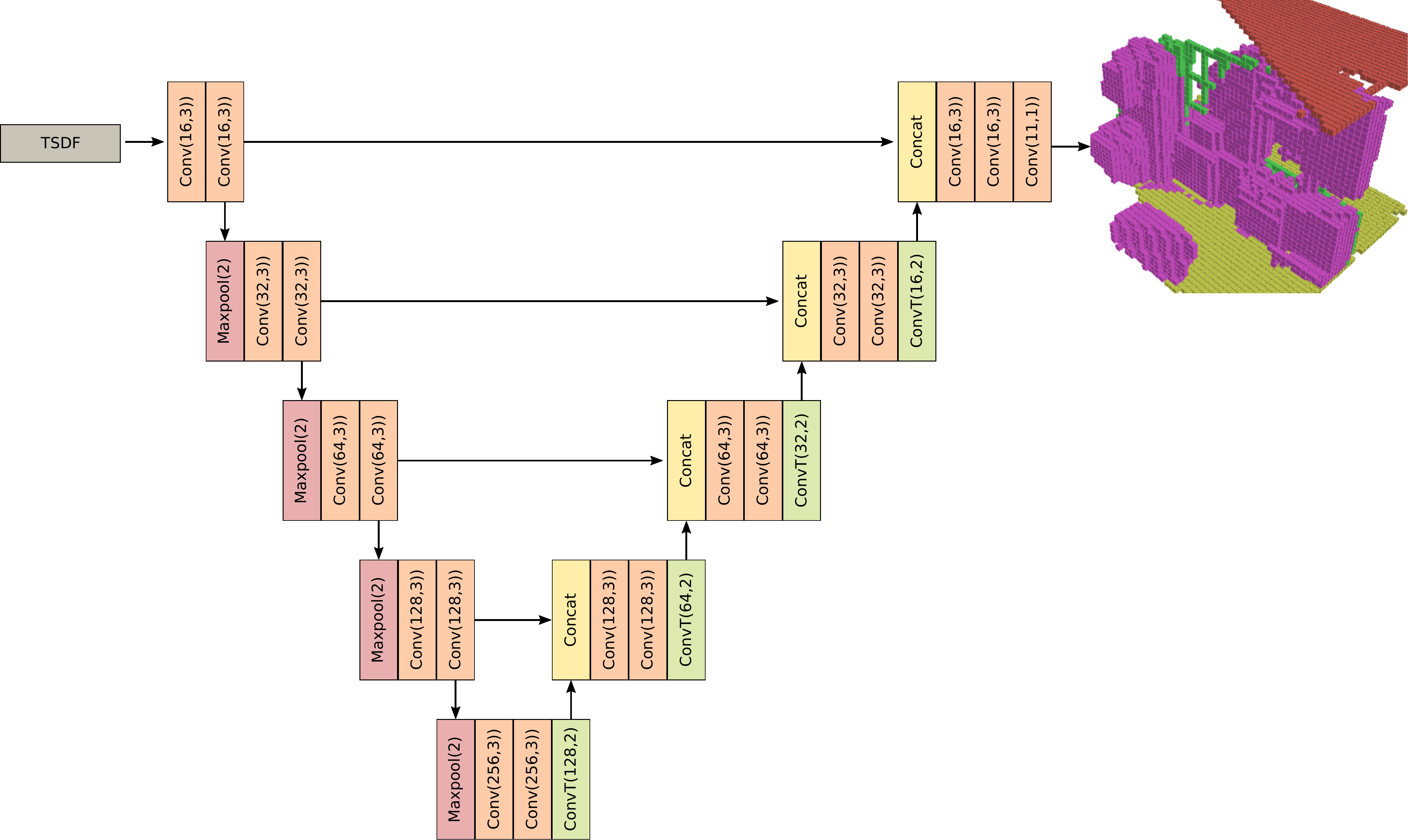}%
\label{fig:UNet_arch}}
\caption{Architecture of the BCNNs used for MNIST and SUNCG experiments.
Conv(d, k, l) stands for a 3D convolution filter stack of depth d and kernel size k and dilation l.
ConvT(d, k) is the up sampling operation Transposed Convolution with depth d, kernel and stride k.
Batch normalization and softplus activation is performed after every Conv layer. Softmax in the final layer.}
\label{fig:arch}
\end{figure*}

We have explored two network architectures.
The first network architecture is inspired by \cite{song2016ssc}, but we have included batch normalization \cite{ioffe2015batch} and
used dilated convolutions instead of max pooling layers and strided convolutions. This to keep the resolution  \cite{springenberg2014striving}. We denote it as SSC-Net.
The second architecture is a UNet \cite{ronneberger2015unet} with max pooling and transposed convolutions as up-sampling.
Just as in \cite{shridhar2019comprehensive} we chose softplus as activation functions instead of relu to have more active weights in the network.
This spread is beneficial for a well calibrated uncertainty \cite{seedat2019calibrated}.
The architecture of the network is displayed in Figure \ref{fig:arch}.

\section{Training}
For training we use the Adam optimizer \cite{kingma2014adam} with constant learning rate of $10^{-3}$.
We do not use weight decay since we regularize using KL divergence.
For the weights we sample the mean uniformly as $\mu \sim U(-\frac{1}{\sqrt{k^d}}, \frac{1}{\sqrt{k^d}})$,
where $k$ is the kernel size in one dimension and $d$ is the number of dimensions. The parameter $\sigma$ is simply initialized as a chosen constant $\sigma_0$.

As noted in \cite{shridhar2019comprehensive} it is beneficial to set $\sigma_0$ larger than the variance of the prior.
In our own experiments on MNIST we see that this improves the separation in predictive entropy. See the experiments section for details.
\section{Evaluation}
To measure the performance of the network we use the Intersection over Union (IoU), which is standard for segmentation tasks.
The main goal is to compare the Bayesian implementation with the deterministic and see that the score is similar.
Just like \cite{song2016ssc} and most works we don't evaluate voxels that are outside of the camera field of view.
Unlike previous works we do evaluate on pixels outside of the room so that the network can learn to predict a room layout
e.g. even though the room in just glimpsed through a door.
\\

To evaluate if the Bayesian approach gives us more information about model uncertainty we look at the scores and uncertainty metrics for
true positives (TP) and false negatives (FN) for each class or for different distances to the surfaces.
If the knowledge about the model limitations is good, there should be a good separation in scores and/or uncertainty for TP and FN.
We measure this with the Bhattacharyya coefficient (BC) \cite{bhattacharyya}, which is an approximation of the amount of overlap of two distributions.
The data is split into $N$ partitions, where in each partition we count the number of TP $q_i$ and FN $p_i$ and calculate the BC as
\begin{equation*}
  BC(\mathbf{p},\mathbf{q}) = \frac{1}{N} \sum_{i=1}^N \sqrt{p_i q_i}.
\end{equation*}
\\

Finally, we also compute the Mean Average Precision (mAP) as the area under the Precision-Recall curve.
This measures both the separation and the accuracy of the model.
We use 101 thresholds $T = \{0, 0.01, ..., 1.0\}$, just as the detection challenge COCO \cite{COCO}. %PASCAL VOC \cite{VOC}
We follow their implementation and use interpolated precision to get a smoother curve.

\section{MNIST experiments}
For MNIST we created a simple 2D CNN using the same building blocks as we use in the 3D SSC-Net.
We used the same code base and tools to allow for faster debugging and experiments.
The network is trained in both Bayesian and Deterministic mode.
\subsection{Baseline experiment}
Here we have trained the network on the training set and evaluated on the test set.
Table \ref{tb:MNIST_AP} shows the Accuracy and mAP for each mode.
We see that the Deterministic CNN without weight decay, i.e. $\omega=0$, performs best.
Our Bayesian version is close behind with the benefit of having built in regularization in terms of a prior.
By explicitly adding regularization in form of weight decay on the deterministic CNN we get the worst performance, especially noticeable in mAP.

\begin{table}[!t]
\renewcommand{\arraystretch}{1.2}
% if using array.sty, it might be a good idea to tweak the value of
% \extrarowheight as needed to properly center the text within the cells
\caption{Compares Accuracy and mAP on MNIST for the deterministic and Bayesian mode of our simple network.}
\label{tb:MNIST_AP}
\centering
% Some packages, such as MDW tools, offer better commands for making tables
% than the plain LaTeX2e tabular which is used here.
\begin{tabular}{|c|c|c|}
\hline
\textbf{Mode} & \textbf{Accuracy}& \textbf{mAP}\\
\hline
Bayesian, Cauchy(0.3), $\sigma_0=0.5$ & 97\% & 0.96\\
\hline
Deterministic, $\omega=0$ & \textbf{98}\% & \textbf{0.97}\\
\hline
Deterministic, $\omega=0.01$ & 97\% & 0.95\\
\hline
\end{tabular}
\end{table}
\subsection{Leave one out experiment}
To see how well the uncertainty measurements captured model uncertainty we conducted an experiment
where digit $0$ was left out from the training data and then introduced at test time.
Unless stated otherwise the Bayesian model uses Cauchy(0.3) as prior and $\sigma_0=0.5$ as initialization.
\subsubsection{Prior and Initialization}
\begin{figure*}[!t]
\centering
\includegraphics[width=\textwidth]{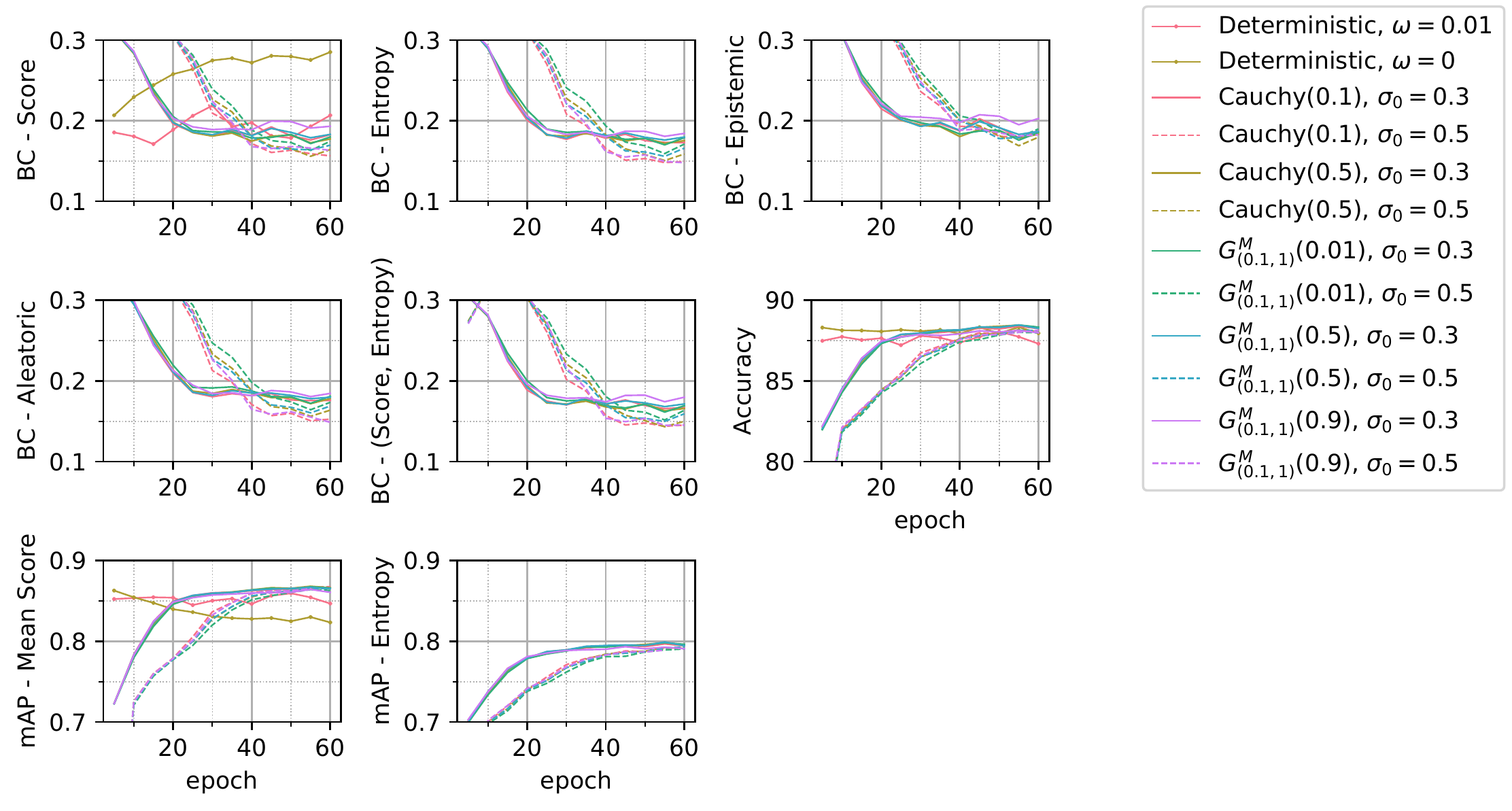}
\caption{BC, mAP and accuracy for different configurations of prior and $\sigma_0$ for the MNIST experiment where digit $0$ is introduced at test time.
We see that high $\sigma_0$ seems to improve BC but increase training time and a prior which is more narrow than $\sigma_0$ gets a lower BC score.
We also see that Entropy, Aleatoric uncertainty and Entropy coupled with score seems to have best separation.
The Deterministic version are comparable in either mAP or BC, but not both at the same time.}
\label{fig:MNIST_prior}
\end{figure*}

In Figure \ref{fig:MNIST_prior} we see how the BC and mAP changes for different priors and $\sigma_0$ during training.
A larger $\sigma_0$ decreases the BC score (increase separation) but require longer training times.
By binning the results on both Entropy and mean score we get the best separation, which indicates that
we can most likely train a classifier on these metrics and do better than the mean score.
This is analogous with the GPOM \cite{gpom} mentioned in section \ref{sec:related}.

\subsubsection{Activation function}
\begin{figure*}[!t]
\centering
\includegraphics[width=\textwidth]{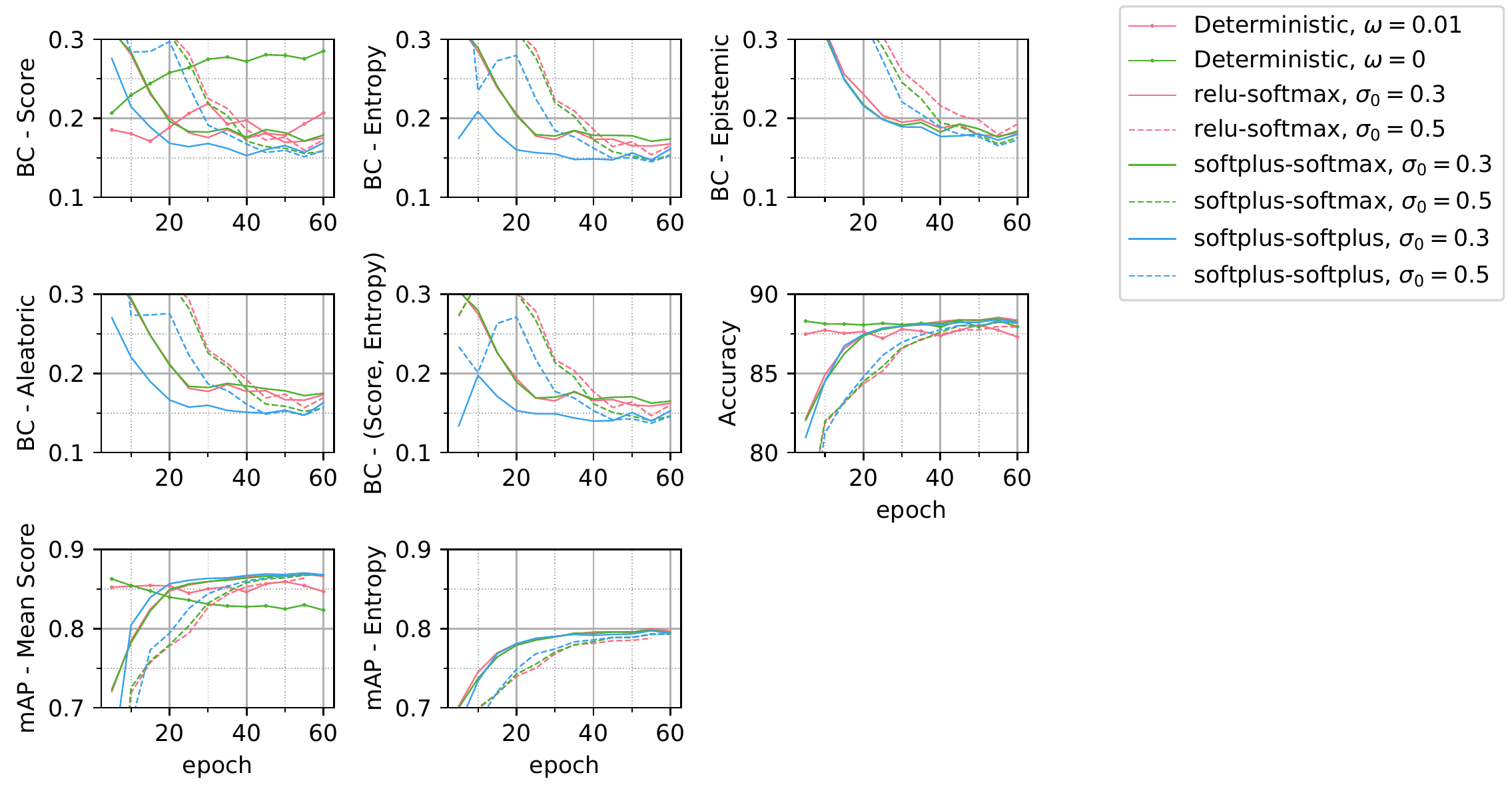}
\caption{BC, mAP and accuracy for different activation functions and $\sigma_0$ for the MNIST experiment where digit $0$ is introduced at test time.
We observe that softplus as a layer activation function in general gets a lower BC score than ReLU.
For the final activation the normalized softplus reaches a lower BC faster than softmax.}
\label{fig:MNIST_activation}
\end{figure*}

Figure \ref{fig:MNIST_activation} shows how the choice of activation function affects the result.
Just as \cite{shridhar2019comprehensive} we observe that softplus in general yields better separation,
both when used as layer activation and normalized as final activation.

\subsubsection{Output distributions}
\begin{figure*}[!t]
\centering
\subfloat[Deterministic]{\includegraphics[width=.38\textwidth]{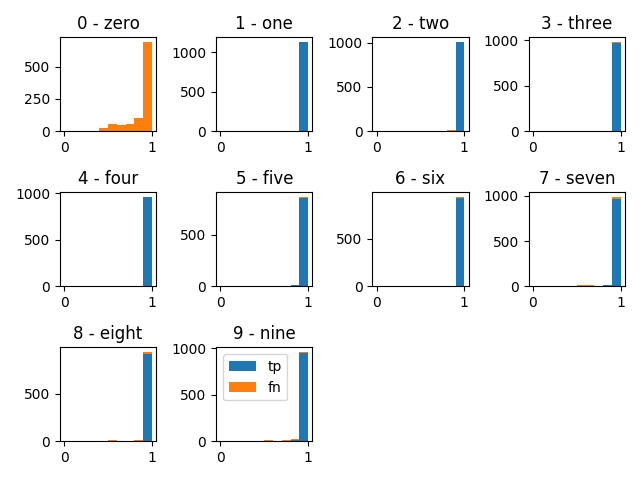}%
\label{fig:MNIST_deterministic}}
\hfil
\subfloat[Deterministic Regularized]{\includegraphics[width=.38\textwidth]{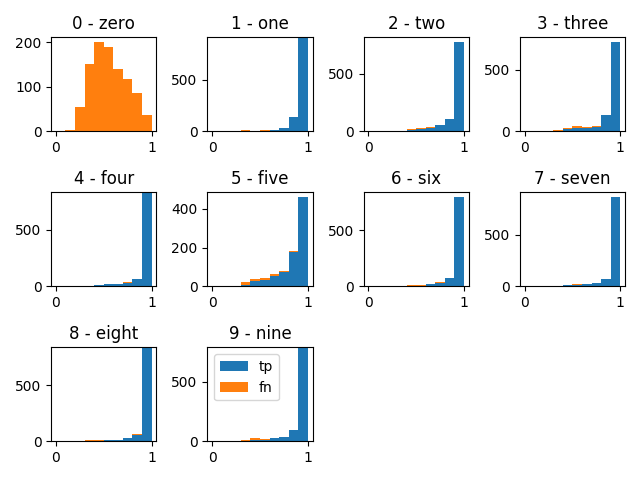}%
\label{fig:MNIST_deterministic_reg}}
\hfil
\subfloat[Bayesian Mean Score]{\includegraphics[width=.38\textwidth]{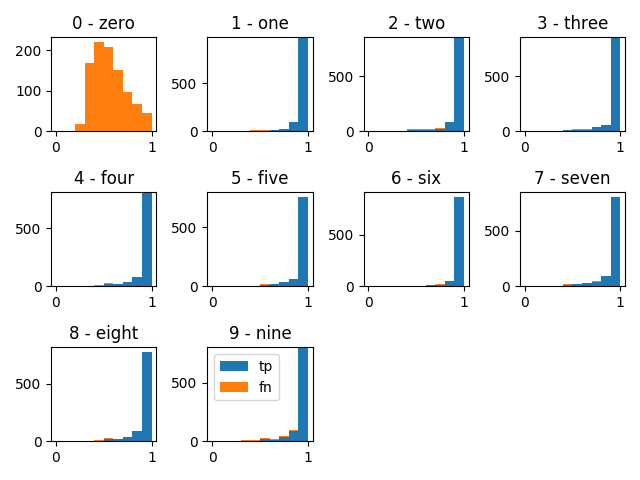}%
\label{fig:MNIST_bayesian}}
\hfil
\subfloat[Bayesian Entropy]{\includegraphics[width=.38\textwidth]{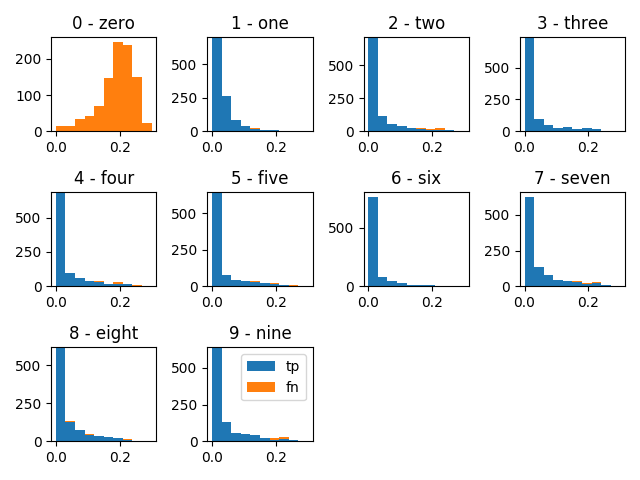}%
\label{fig:MNIST_entropy}}
\caption{Here we see true and false predictions for all digits in the test set when $0$ has been left out from training.
Blue bars are number of true predictions with the binned score, while orange are false predictions.
In \ref{fig:MNIST_deterministic} we have the deterministic CNN which has high belief in many false $0$ predictions.
In \ref{fig:MNIST_deterministic_reg} we have the deterministic CNN with regularization $\omega=0.01$ which has a more balanced belief, but overall lower.
In \ref{fig:MNIST_bayesian} and \ref{fig:MNIST_entropy} we see lower certainty for digit $0$ but still high certainty for other classes in both score and entropy respectively.}
\label{fig:MNIST_hist}
\end{figure*}

For a more in depth understanding of the output we also look at the output distributions.
As described in section \ref{sec:prediction} we have calculated the predictive mean $\bar{\pt}_i$ and entropy $H_i$ for each sample indexed by $i$ in the dataset.
As usual we get the predicted label index as $l_i = \argmax{\bar{\pt}_i}$ with corresponding score $\bar{\pt}_i(l_i)$ and entropy $H_i(l_i)$.
In Figure \ref{fig:MNIST_hist} we form histograms of these for all samples in the test dataset categorized by their true class.
For the deterministic case we see that the regularization helps separate the true zeros from the other predictions.
However, the distributions for the other classes are much wider, indicating that the network is more uncertain overall.
In the Bayesian case we see that the distribution for true zeros are well separated from the others, while maintaining certainty for the other classes.
The entropy also seems to be a good indicator.

%
%MNIST/Apr14_17-23-56_sweep/prior:{name:cauchy,gamma:0.3}|var_init:0.5|/val, 60 epochs
%
\section{SUNCG mini experiments}
SUNCG \cite{song2016ssc} is a large dataset consisting of 45,622 manually created synthetical indoor scenes with 84 labeled categories.
We've used a subset of 2000 training scenes and 1000 testing scenes for the experiments.
As in \cite{song2016ssc} we've used the flipped Truncated Signed Distance Function (TSDF) as input to our network
and the same 11 categories as output.
We chose the projective version since it is more realistic in a robotic setting.
We've also chosen a coarser grid with 0.08m resolution and 60x40x60 in size which is similar to their output size and resolution.
\subsection{Baseline experiment}
In Table \ref{tb:SUNCG} we see the mean IoU (mIoU) over all categories for the Deterministic and Bayesian CNN.
In general they have similar mIoU but the Bayesian versions have better mAP.
\begin{table}%[!t]
\renewcommand{\arraystretch}{1.2}
% if using array.sty, it might be a good idea to tweak the value of
% \extrarowheight as needed to properly center the text within the cells
\caption{Compares mIoU on SUNCG for the (D)eterministic and (B)ayesian versions of our CNNs.}
\label{tb:SUNCG}
\centering
% Some packages, such as MDW tools, offer better commands for making tables
% than the plain LaTeX2e tabular which is used here.
\begin{tabular}{|c|c|c|c|c|}
\hline
\textbf{Type} & \textbf{Arch.} & \textbf{Parameters} &\textbf{mIoU} & \textbf{mAP}\\
\hline
%ICPR2_eval/suncg11|net:bssc|train:data3|val:data3/Jun30_12-32-17 - epoch 150
B & SSC-Net & Cauchy(0.05), $\sigma_0$=0.05, $\beta$=0.1 & 0.22 & 0.26\\
\hline
%ICPR2/suncg11_bssc_unet_data3/Jul10_03-48-22_sweep/kl_beta:5|prior:{name:cauchy,gamma:0.1}|var_init:0.1| - epoch 390
B & UNet & Cauchy(0.1), $\sigma_0$=0.1, $\beta$=5 & \textbf{0.23} & \textbf{0.29}\\
\hline
%ICPR2_eval/suncg11|net:det_ssc-wd:0|train:data3|val:data3/Jun30_09-45-35 - epoch 30
D & SSC-Net & $\omega$=0 & \textbf{0.23} & 0.24\\
\hline
%CPR2/suncg11_det_unet_data3/Jul12_12-21-29_sweep/weight_decay:0 - epoch 30
D & UNet & $\omega$=0 & 0.21 & 0.21\\
\hline
%ICPR2/suncg11|net:det_ssc-wd:0.01|train:data3|val:data3/Jul04_23-07-51 - epoch 30
D & SSC-Net & $\omega$=0.01 & 0.15 & 0.19\\
\hline
%CPR2/suncg11_det_unet_data3/Jul12_12-21-29_sweep/weight_decay:0.01 - epoch 190
D & UNet & $\omega$=0.01 & 0.16 & 0.20\\
\hline
\end{tabular}
\end{table}
\subsection{Complexity cost weight $\beta$}
\begin{figure*}[!t]
\centering
\includegraphics[width=\textwidth]{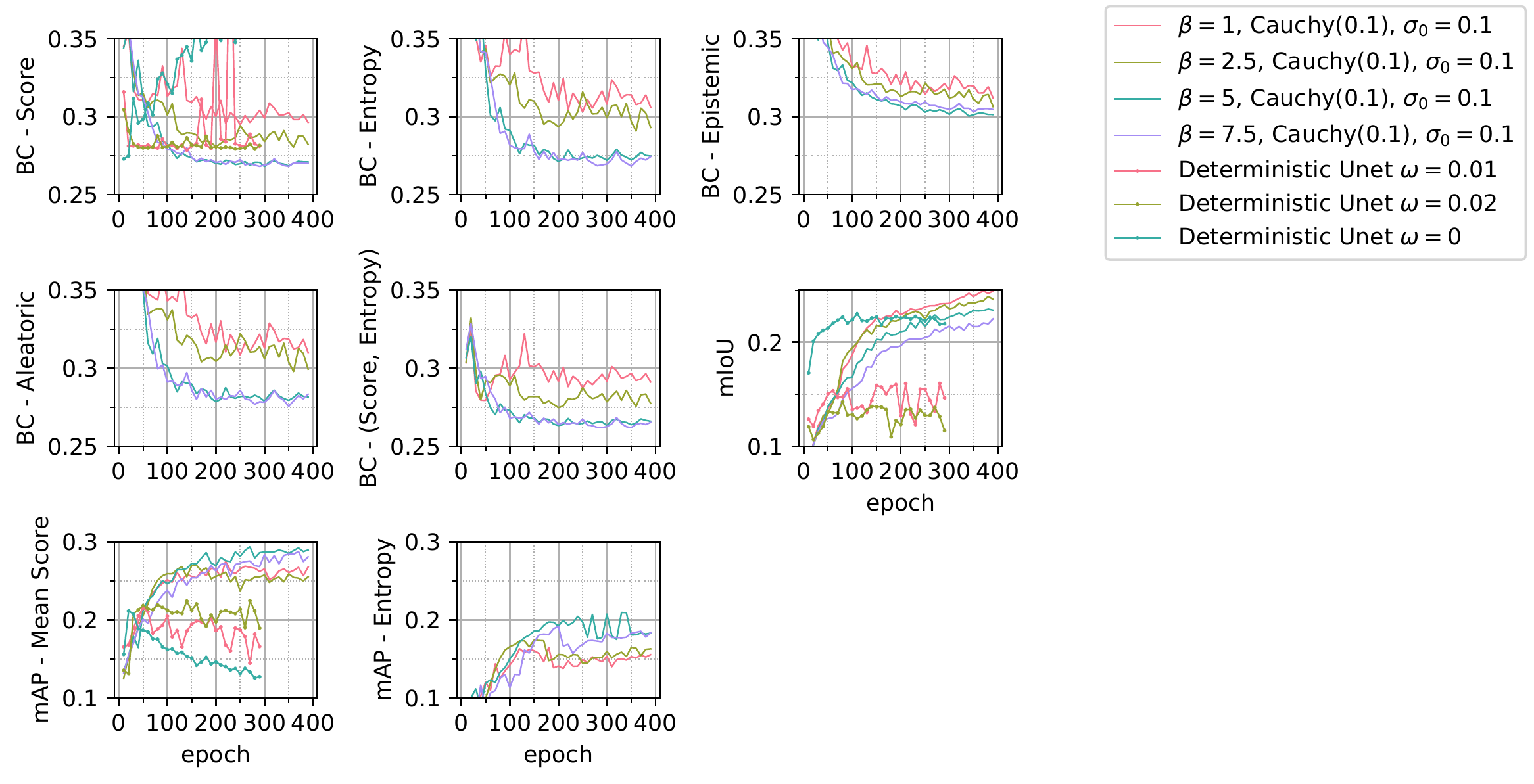}
\caption{BC, mAP and mIoU for the Bayesian UNet with different weights $\beta$ and $\omega$ for the SUNCG mini dataset.
We observe that $\beta=5$ is better in all metrics but mIoU, where $\beta=1$ is best.}
\label{fig:SUNCG_beta}
\end{figure*}

For the Bayesian UNet we conducted a $\beta$ parameter experiment on SUNCG, which weighs the complexity cost against the likelihood cost.
As a reference, the deterministic version with different weight decays are also included.
We see in Figure \ref{fig:SUNCG_beta} that $\beta=5$ is better in all metrics but mIoU, where $\beta=1$ has the highest score.
We also see that all Bayesian versions are better than the deterministic in mAP and
that a too large $\beta$ makes the model unable to fit the data properly, while too small will get a better fit but worse separation.
\subsection{Leave one out} \label{sec:SUNCG_leave_one_out}
For the SUNCG mini dataset we removed all instances of the class \textit{bed} during training and then tested on the full dataset.
The result is presented in Table \ref{tb:SUNCG_nobed}.
Here we test both the SSC-Net and the UNet architecture in both Bayesian and Deterministic versions.
We observe that both Bayesian versions outperform the Deterministic in mIoU, mAP and BC.
For the sample based metrics the Bayesian SSC-Net outperform the Bayesian UNet, but this might not be the case given more data.
\begin{table}%[!t]
\renewcommand{\arraystretch}{1.2}
\caption{BC, mAP and mIoU for different network architectures when the \textit{bed} class is removed from training.
S=Score, E=Entropy.
We observe that Bayesian SSC-Net has the best score in most metrics.}
\label{tb:SUNCG_nobed}
\centering
\begin{tabular}{|c|c|c|c|c|c|}
\hline
\textbf{CNN} & \textbf{mIoU} & \textbf{mAP: S} & \textbf{mAP: E} & \textbf{BC: S} & \textbf{BC: E}\\
\hline
SSC-Net $\omega$=0 & 0.19 & 0.2 &  & 0.31 & \\
\hline
SSC-Net $\omega$=0.01 & 0.14 & 0.23 &  & 0.29 & \\
\hline
UNet $\omega$=0 & 0.2 & 0.14 &  & 0.39 & \\
\hline
UNet $\omega$=0.01 & 0.15 & 0.21 &  & 0.28 & \\
\hline
B-SSC-Net & 0.21 & \textbf{0.26} & \textbf{0.19} & \textbf{0.27} & \textbf{0.28}\\
\hline
B-UNet & \textbf{0.21} & 0.25 & 0.17 & 0.28 & 0.28\\
\hline
\end{tabular}

\end{table}
\subsection{Output distributions}
\begin{figure}%[!t]
\centering
\subfloat[Bayesian score]{\includegraphics[width=.5\linewidth]{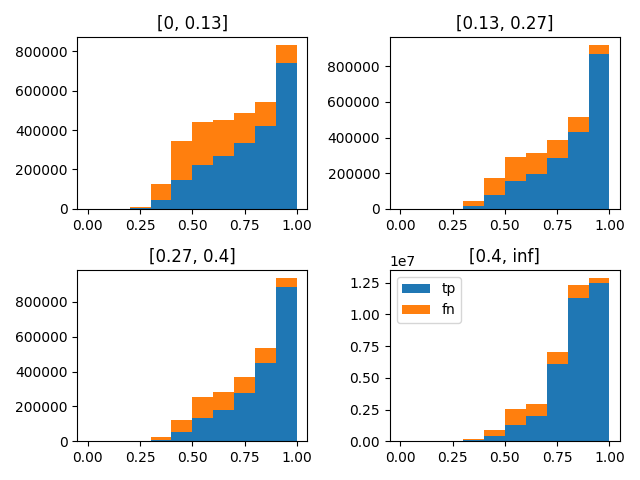}%
\label{fig:SUNCG_tsdf_bayesian}}
\hfil
\subfloat[Bayesian Entropy]{\includegraphics[width=.5\linewidth]{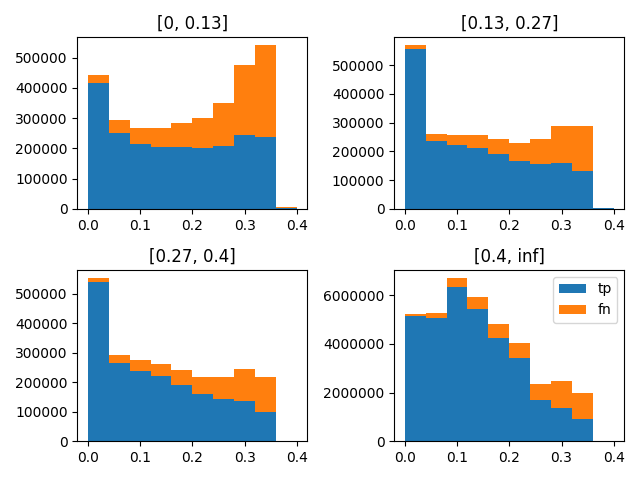}%
\label{fig:SUNCG_tsdf_entropy}}
\caption{Here we see true and false predictions for all voxels at different distances (in meter) from observed surfaces.
In \ref{fig:SUNCG_tsdf_entropy} we see that as the entropy (uncertainty) grows the ratio of false to true predictions increases.
This hold for all areas, but especially close to the observed surfaces. }
\label{fig:SUNCG_tsdf_hist}
\end{figure}

In Figure \ref{fig:SUNCG_tsdf_hist} we see histograms for true and false predictions at different distances from the observed surface
for the Bayesian SSC-Net.
We see that as the entropy (uncertainty) grows the ratio of false to true predictions increases, this is what we would expect from a sound
measurement of uncertainty. It seems to have a better grasp of the uncertainty close to the observed surface.
\subsection{Example output}
See Figure \ref{fig:SUNCG_example} for example output.
More examples are available in the supplementary material.
\begin{figure*}[!t]
\captionsetup[subfigure]{labelformat=empty}
\centering
\subfloat[]{\includegraphics[width=.6\textwidth]{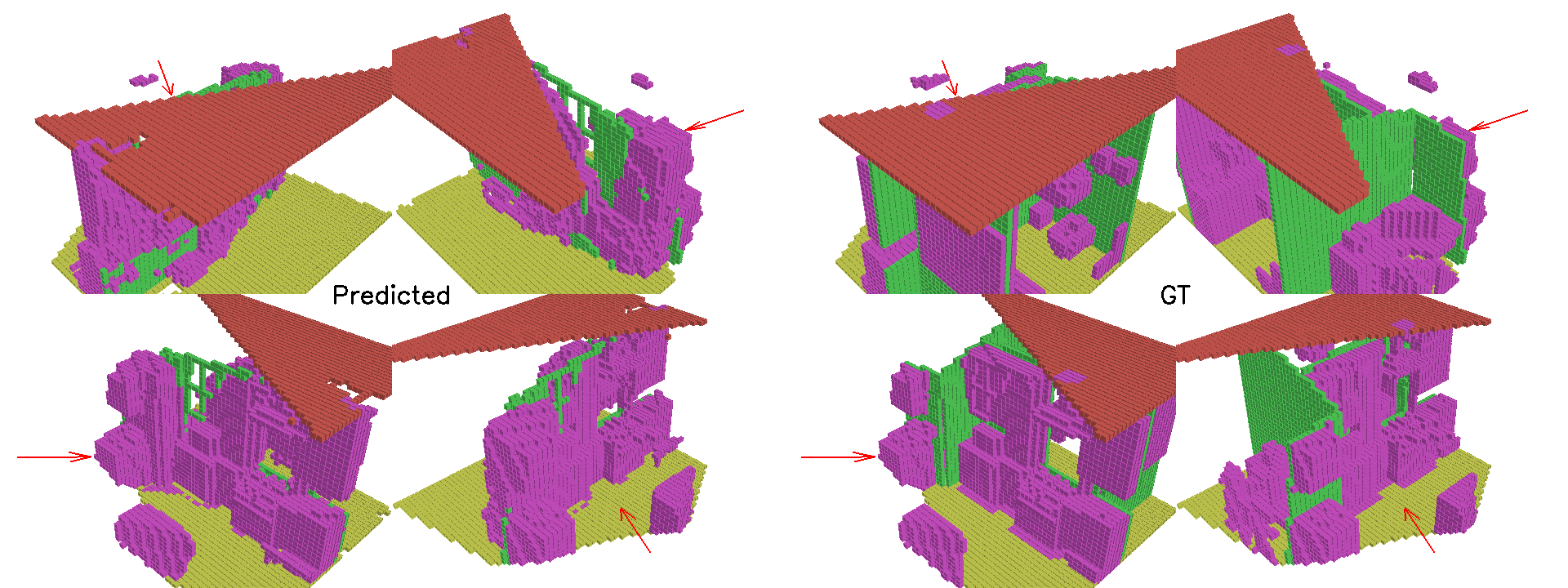}}%
\hfil
\subfloat[]{\includegraphics[width=.3\textwidth, trim=0 0 1000 0, clip]{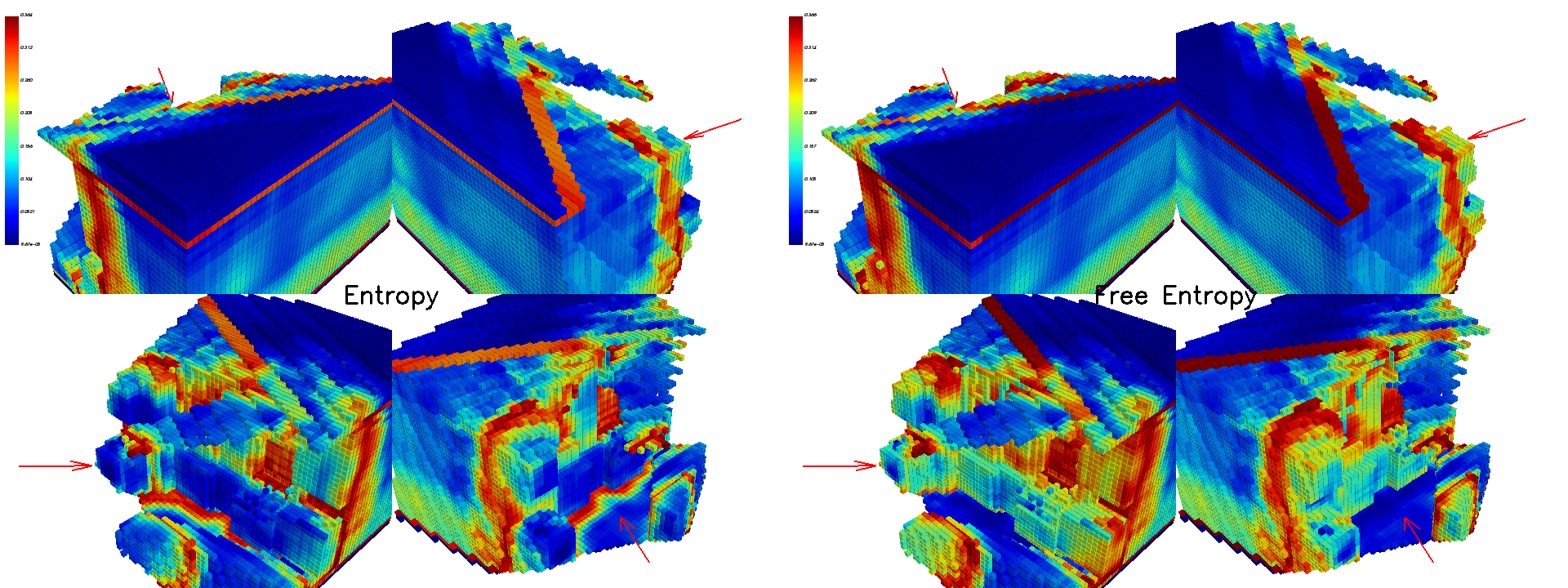}}%
\caption{Here is one example output from the SUNCG mini test set. From the left we have predicted labels, ground truth and entropy.}
\label{fig:SUNCG_example}
\end{figure*}

\section{Conclusion}
We've seen that a Bayesian CNN is clearly better at being uncertain when presented with unseen data.
This was shown with a simple CNN model on the MNIST dataset.
In the same experiment we also showed that the best separation between true and false predictions is found when combining entropy and mean score.
This means that a final classifier layer could be added to get better probability estimates.
\\

For the Semantic Scene Completion task we show that the Bayesian CNNs outperfom the Deterministic CNNs
when faced with a category not seen at training time. This is shown by evaluating mIoU, mAP and separation score BC.
Due to the complexity of the task and the model capacity we cannot show the same well separated output distributions as in the MNIST experiment. We do show that entropy represents the uncertainty well, especially close to the observed surface.
\\

Future work will be to use an architecture better suited for sparse data and to utilize the RGB information from the image as well.
We also want to look at how the uncertainty scores can be used in navigation.
\section*{Acknowledgment}
This work is supported by Wallenberg AI Autonomous Systems and Software Program (WASP).

\IEEEtriggeratref{6}


\bibliographystyle{IEEEtran}
% \clearpage
% \pagebreak
\bibliography{IEEEabrv,refs}

\begin{thebibliography}{10}
\providecommand{\url}[1]{#1}
\csname url@samestyle\endcsname
\providecommand{\newblock}{\relax}
\providecommand{\bibinfo}[2]{#2}
\providecommand{\BIBentrySTDinterwordspacing}{\spaceskip=0pt\relax}
\providecommand{\BIBentryALTinterwordstretchfactor}{4}
\providecommand{\BIBentryALTinterwordspacing}{\spaceskip=\fontdimen2\font plus
\BIBentryALTinterwordstretchfactor\fontdimen3\font minus
  \fontdimen4\font\relax}
\providecommand{\BIBforeignlanguage}[2]{{%
\expandafter\ifx\csname l@#1\endcsname\relax
\typeout{** WARNING: IEEEtran.bst: No hyphenation pattern has been}%
\typeout{** loaded for the language `#1'. Using the pattern for}%
\typeout{** the default language instead.}%
\else
\language=\csname l@#1\endcsname
\fi
#2}}
\providecommand{\BIBdecl}{\relax}
\BIBdecl

\bibitem{song2016ssc}
S.~Song, F.~Yu, A.~Zeng, A.~X. Chang, M.~Savva, and T.~Funkhouser, ``Semantic
  scene completion from a single depth image,'' \emph{Proceedings of 30th IEEE
  Conference on Computer Vision and Pattern Recognition}, 2017.

\bibitem{Garbade_2019_CVPR_Workshops}
M.~Garbade, Y.-T. Chen, J.~Sawatzky, and J.~Gall, ``Two stream 3d semantic
  scene completion,'' in \emph{The IEEE Conference on Computer Vision and
  Pattern Recognition (CVPR) Workshops}, June 2019.

\bibitem{Liu_NIPS2018}
\BIBentryALTinterwordspacing
S.~Liu, Y.~HU, Y.~Zeng, Q.~Tang, B.~Jin, Y.~Han, and X.~Li, ``See and think:
  Disentangling semantic scene completion,'' in \emph{Advances in Neural
  Information Processing Systems 31}, S.~Bengio, H.~Wallach, H.~Larochelle,
  K.~Grauman, N.~Cesa-Bianchi, and R.~Garnett, Eds.\hskip 1em plus 0.5em minus
  0.4em\relax Curran Associates, Inc., 2018, pp. 263--274. [Online]. Available:
  \url{http://papers.nips.cc/paper/7310-see-and-think-disentangling-semantic-scene-completion.pdf}
\BIBentrySTDinterwordspacing

\bibitem{Dai_2018_CVPR}
A.~Dai, D.~Ritchie, M.~Bokeloh, S.~Reed, J.~Sturm, and M.~Nießner,
  ``Scancomplete: Large-scale scene completion and semantic segmentation for 3d
  scans,'' in \emph{The IEEE Conference on Computer Vision and Pattern
  Recognition (CVPR)}, June 2018.

\bibitem{dai20183dmv}
A.~Dai and M.~Nie{\ss}ner, ``3dmv: Joint 3d-multi-view prediction for 3d
  semantic scene segmentation,'' in \emph{Proceedings of the European
  Conference on Computer Vision (ECCV)}, 2018, pp. 452--468.

\bibitem{dai2017scannet}
A.~Dai, A.~X. Chang, M.~Savva, M.~Halber, T.~Funkhouser, and M.~Nie{\ss}ner,
  ``Scannet: Richly-annotated 3d reconstructions of indoor scenes,'' in
  \emph{Proc. Computer Vision and Pattern Recognition (CVPR), IEEE}, 2017.

\bibitem{SparseConvNet_CVPR2018}
B.~Graham, M.~Engelcke, and L.~van~der Maaten, ``3d semantic segmentation with
  submanifold sparse convolutional networks,'' \emph{CVPR}, 2018.

\bibitem{thomas2019KPConv}
H.~Thomas, C.~R. Qi, J.-E. Deschaud, B.~Marcotegui, F.~Goulette, and L.~J.
  Guibas, ``Kpconv: Flexible and deformable convolution for point clouds,''
  \emph{Proceedings of the IEEE International Conference on Computer Vision},
  2019.

\bibitem{wu2018pointconv}
W.~Wu, Z.~Qi, and L.~Fuxin, ``Pointconv: Deep convolutional networks on 3d
  point clouds,'' 2018.

\bibitem{Choy_CVPR2019}
C.~{Choy}, J.~{Gwak}, and S.~{Savarese}, ``4d spatio-temporal convnets:
  Minkowski convolutional neural networks,'' in \emph{2019 IEEE/CVF Conference
  on Computer Vision and Pattern Recognition (CVPR)}, June 2019, pp.
  3070--3079.

\bibitem{blundell2015weight}
C.~Blundell, J.~Cornebise, K.~Kavukcuoglu, and D.~Wierstra, ``Weight
  uncertainty in neural networks,'' \emph{arXiv preprint arXiv:1505.05424},
  2015.

\bibitem{shridhar2019comprehensive}
K.~Shridhar, F.~Laumann, and M.~Liwicki, ``A comprehensive guide to bayesian
  convolutional neural network with variational inference,'' 2019.

\bibitem{gpom}
S.~T. O’Callaghan and F.~T. Ramos, ``Gaussian process occupancy maps,''
  \emph{The International Journal of Robotics Research}, vol.~31, no.~1, pp.
  42--62, 2012.

\bibitem{kuleshov2018accurate}
V.~Kuleshov, N.~Fenner, and S.~Ermon, ``Accurate uncertainties for deep
  learning using calibrated regression,'' \emph{arXiv preprint
  arXiv:1807.00263}, 2018.

\bibitem{seedat2019calibrated}
N.~Seedat and C.~Kanan, ``Towards calibrated and scalable uncertainty
  representations for neural networks,'' 2019.

\bibitem{reparametrization}
M.~Opper and C.~Archambeau, ``The variational gaussian approximation
  revisited,'' \emph{Neural computation}, vol.~21, pp. 786--92, 10 2008.

\bibitem{kendall2017uncertainties}
A.~Kendall and Y.~Gal, ``What uncertainties do we need in bayesian deep
  learning for computer vision?'' in \emph{Advances in neural information
  processing systems}, 2017, pp. 5574--5584.

\bibitem{ioffe2015batch}
S.~Ioffe and C.~Szegedy, ``Batch normalization: Accelerating deep network
  training by reducing internal covariate shift,'' 2015.

\bibitem{springenberg2014striving}
J.~T. Springenberg, A.~Dosovitskiy, T.~Brox, and M.~Riedmiller, ``Striving for
  simplicity: The all convolutional net,'' \emph{arXiv preprint
  arXiv:1412.6806}, 2014.

\bibitem{ronneberger2015unet}
O.~Ronneberger, P.~Fischer, and T.~Brox, ``U-net: Convolutional networks for
  biomedical image segmentation,'' 2015.

\bibitem{kingma2014adam}
D.~P. Kingma and J.~Ba, ``Adam: A method for stochastic optimization,'' 2014.

\bibitem{bhattacharyya}
A.~Bhattacharyya, ``On a measure of divergence between two statistical
  populations defined by their probability distributions,'' \emph{Bull.
  Calcutta Math. Soc.}, vol.~35, pp. 99--109, 1943.

\bibitem{COCO}
T.-Y. Lin, M.~Maire, S.~Belongie, J.~Hays, P.~Perona, D.~Ramanan,
  P.~Doll{\'a}r, and C.~L. Zitnick, ``Microsoft coco: Common objects in
  context,'' in \emph{Computer Vision -- ECCV 2014}, D.~Fleet, T.~Pajdla,
  B.~Schiele, and T.~Tuytelaars, Eds.\hskip 1em plus 0.5em minus 0.4em\relax
  Cham: Springer International Publishing, 2014.

\end{thebibliography}

\clearpage
\pagebreak

\section*{Appendix A. Output distributions SUNCG}
\begin{figure}[h]
\centering
\subfloat[Bayesian SSC-Net entropy]%
{\includegraphics[width=\tsdfwidth]{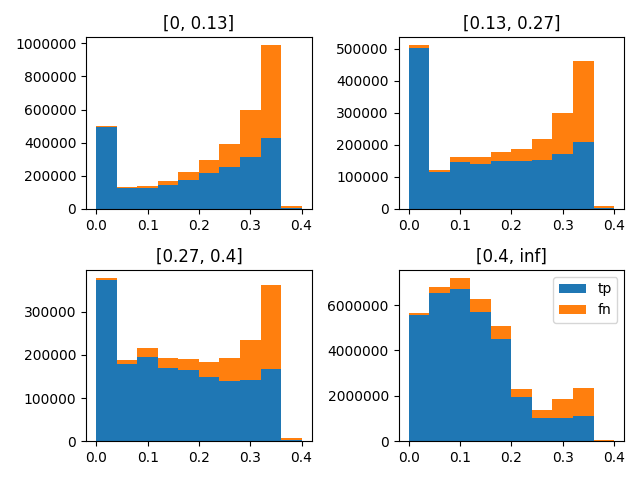}%
\label{fig:SUNCG_tsdf_bssc_entropy}}
\hfil
\subfloat[Bayesian UNet entropy]%
{\includegraphics[width=\tsdfwidth]{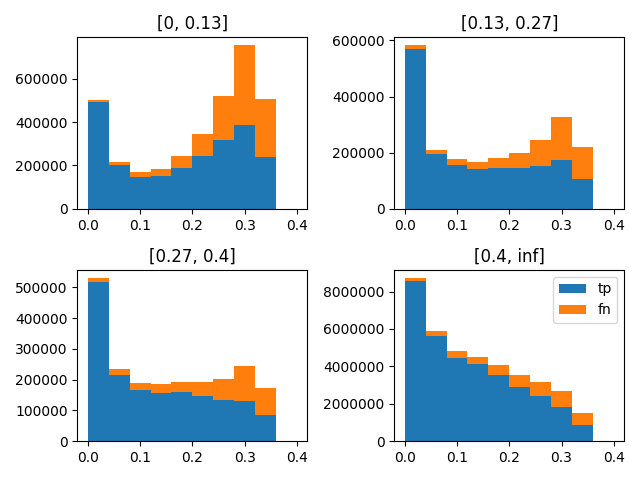}%
\label{fig:SUNCG_tsdf_bunet_entropy}}
\caption{True and false predictions for all voxels at different distances (in meter) from observed surfaces.
We see that as the entropy (uncertainty) grows the ratio of false to true predictions increases.
This hold for all areas, but especially close to the observed surfaces. }
\label{fig:SUNCG_nobed_entropy}
\end{figure}

\begin{figure}[h]
\centering
\subfloat[SSC-Net score, $\omega=0$]%
{\includegraphics[width=\tsdfwidth]{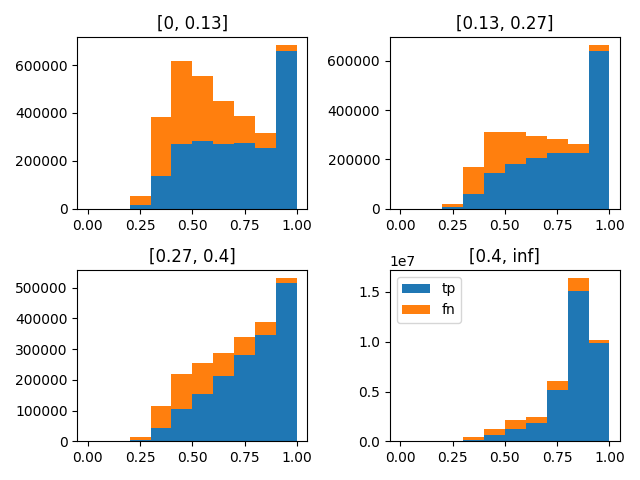}%
\label{fig:SUNCG_tsdf_ssc_det}}
\hfil
\subfloat[SSC-Net score, $\omega=0.01$]%
{\includegraphics[width=\tsdfwidth]{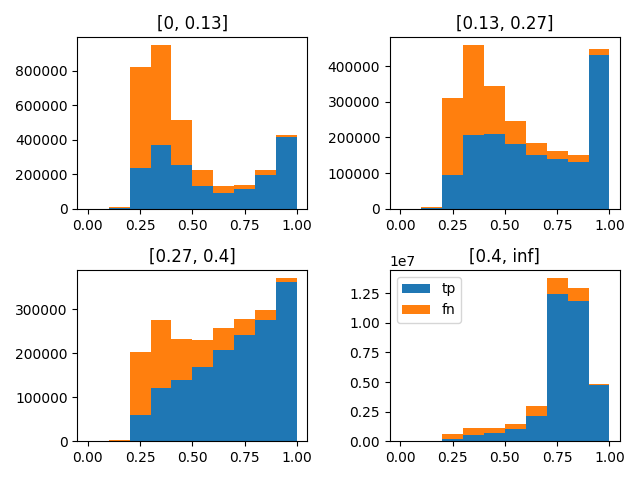}%
\label{fig:SUNCG_tsdf_ssc_det_reg}}
\hfil
\subfloat[UNet score, $\omega=0$]%
{\includegraphics[width=\tsdfwidth]{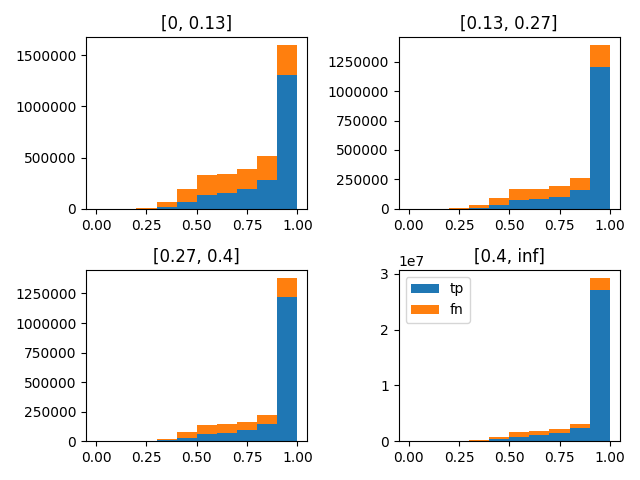}%
\label{fig:SUNCG_tsdf_unet_det}}
\hfil
\subfloat[UNet score, $\omega=0.01$]%
{\includegraphics[width=\tsdfwidth]{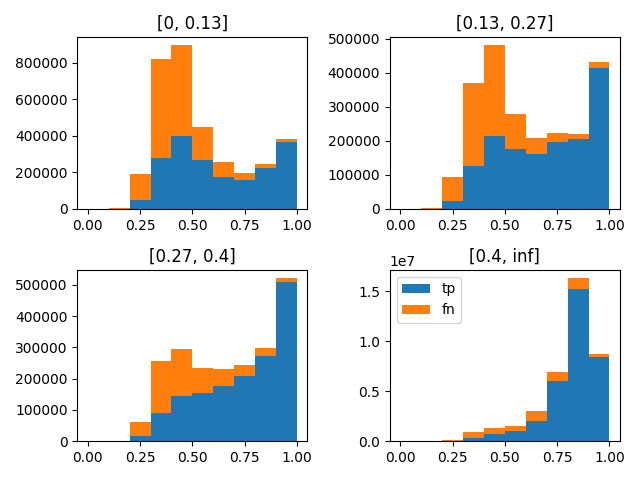}%
\label{fig:SUNCG_tsdf_unet_det_reg}}
\hfil
\subfloat[Bayesian SSC-Net score]%
{\includegraphics[width=\tsdfwidth]{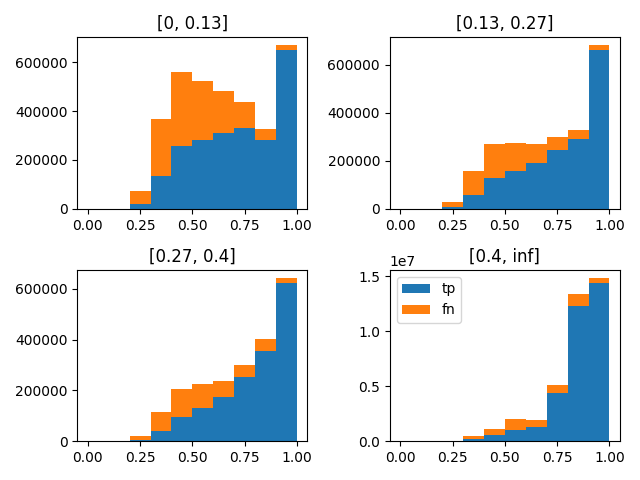}%
\label{fig:SUNCG_tsdf_bssc}}
\hfil
\subfloat[Bayesian UNet score]%
{\includegraphics[width=\tsdfwidth]{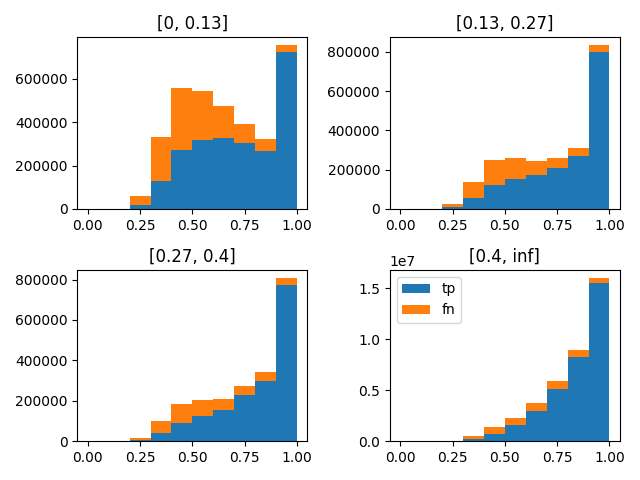}%
\label{fig:SUNCG_tsdf_bunet}}
\caption{True and false predictions for all voxels at different distances (in meter) from observed surfaces.
We see the distribution of scores for all architectures.}
\label{fig:SUNCG_nobed_score}
\end{figure}

In Figure \ref{fig:SUNCG_nobed_entropy} and \ref{fig:SUNCG_nobed_score} are histograms of
output entropy and scores respectively for the 4 different network configurations in the
SUNCG leave-one-out experiment in Section \ref{sec:SUNCG_leave_one_out}.
There are 4 different histograms per network, each bins true (TP) and false positives (FP) with a
certain distance from observed surface.

For the scores in \ref{fig:SUNCG_nobed_score} we see that in general the ratio of FP
to TP grow as score decreases, at least close to the observed surface.
For entropy the ratio grow as entropy increases, which we expect from a measurement of uncertainty.
\section*{Appendix B. Examples}
\begin{figure*}%[!t]
\captionsetup[subfigure]{labelformat=empty}
\centering
\subfloat[]{\includegraphics[width=4in]{SUNCG_mini_ex0}}%
\hfil
\subfloat[]{\includegraphics[width=2in, trim=0 0 1000 0, clip]{SUNCG_mini_ex0_entropy}}%
\vspace{-12pt}
\hfil
\subfloat[]{\includegraphics[width=4in]{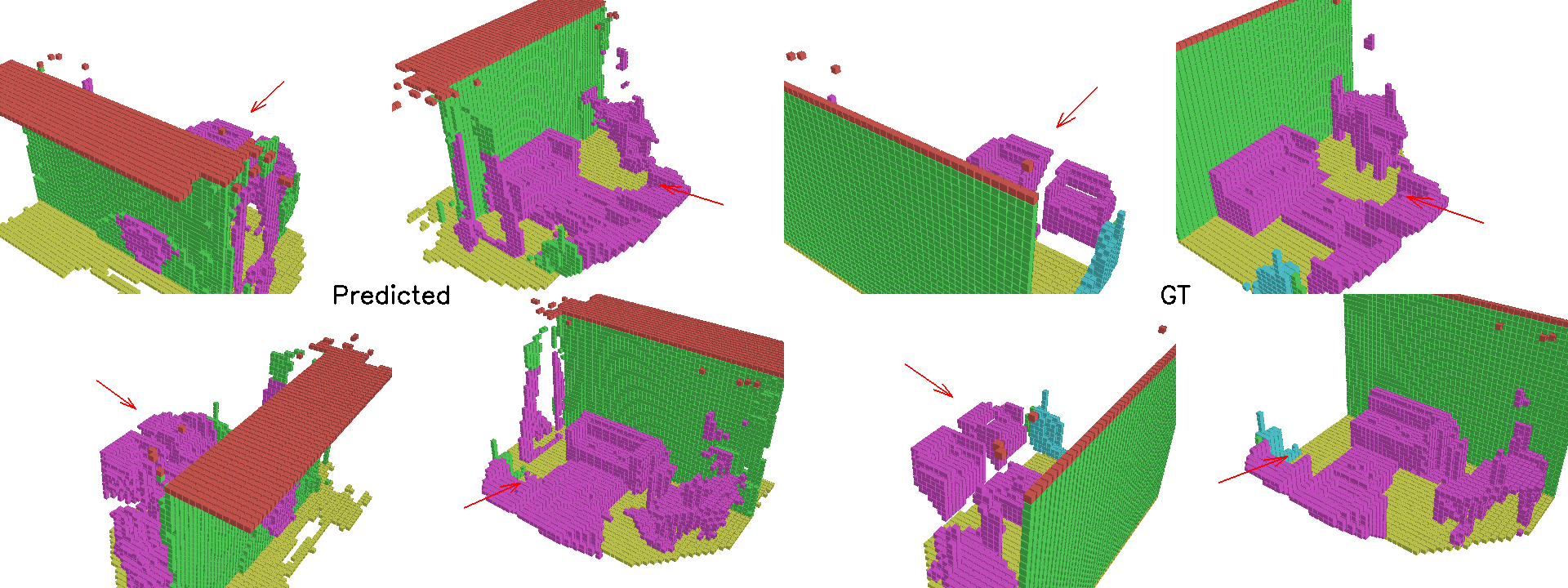}}%
\hfil
\subfloat[]{\includegraphics[width=2in, trim=0 0 1000 0, clip]{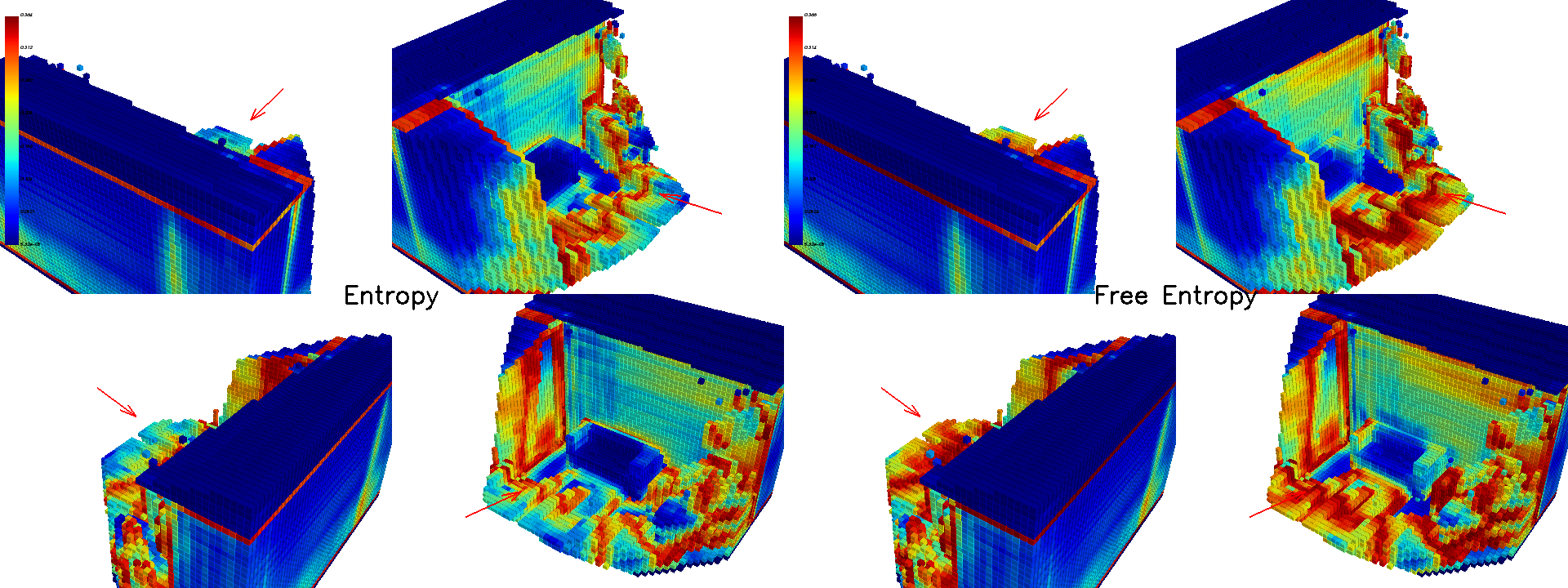}}%
\vspace{-12pt}
\hfil
\subfloat[]{\includegraphics[width=4in]{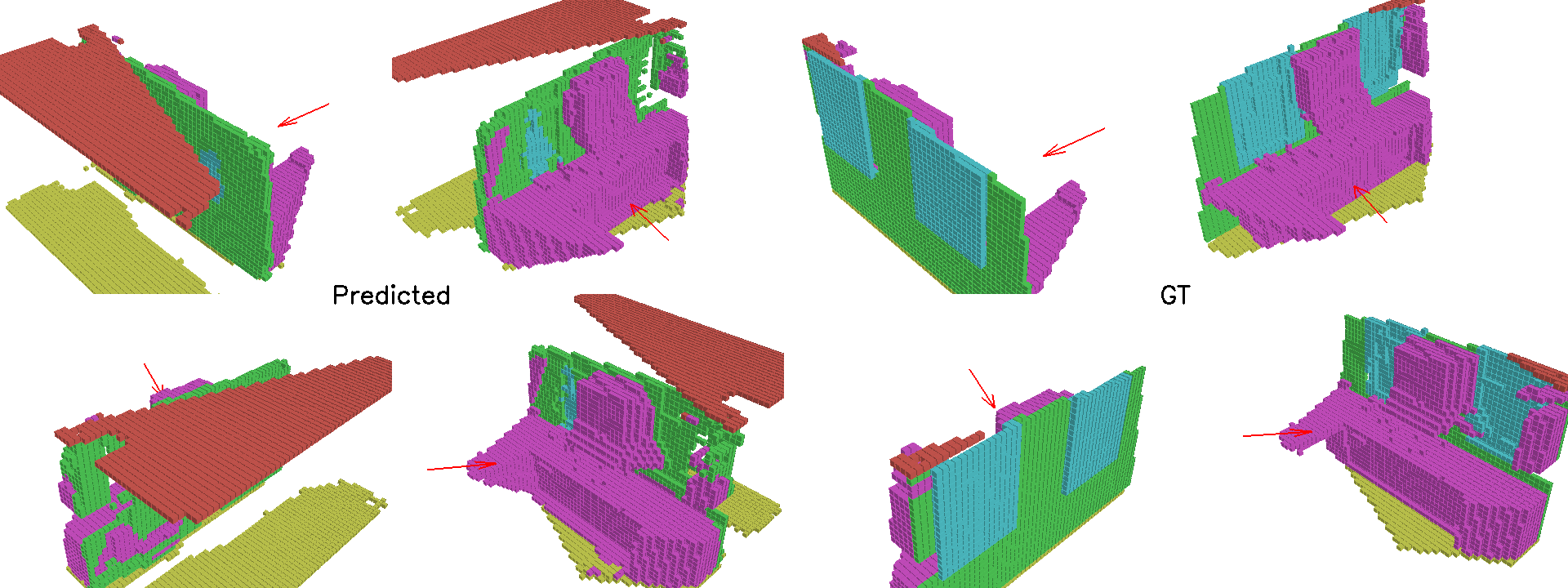}}%
\hfil
\subfloat[]{\includegraphics[width=2in, trim=0 0 1000 0, clip]{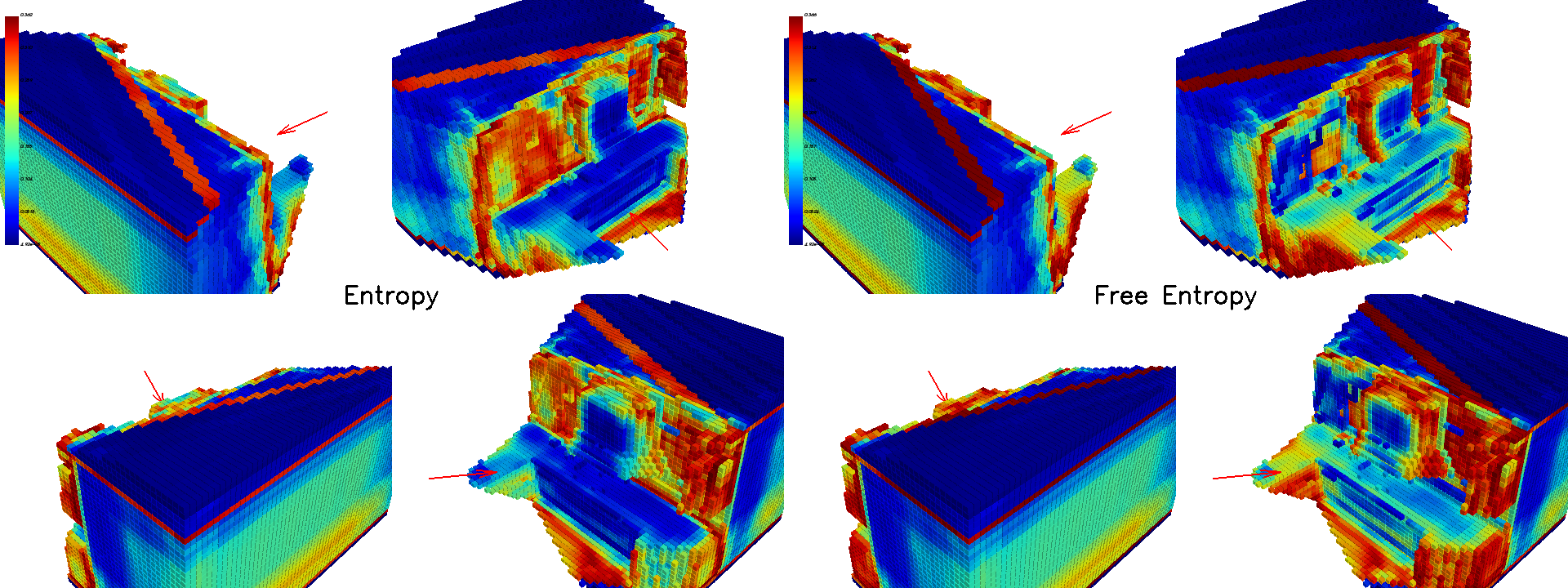}}%
\vspace{-12pt}
\hfil
\subfloat[]{\includegraphics[width=4in]{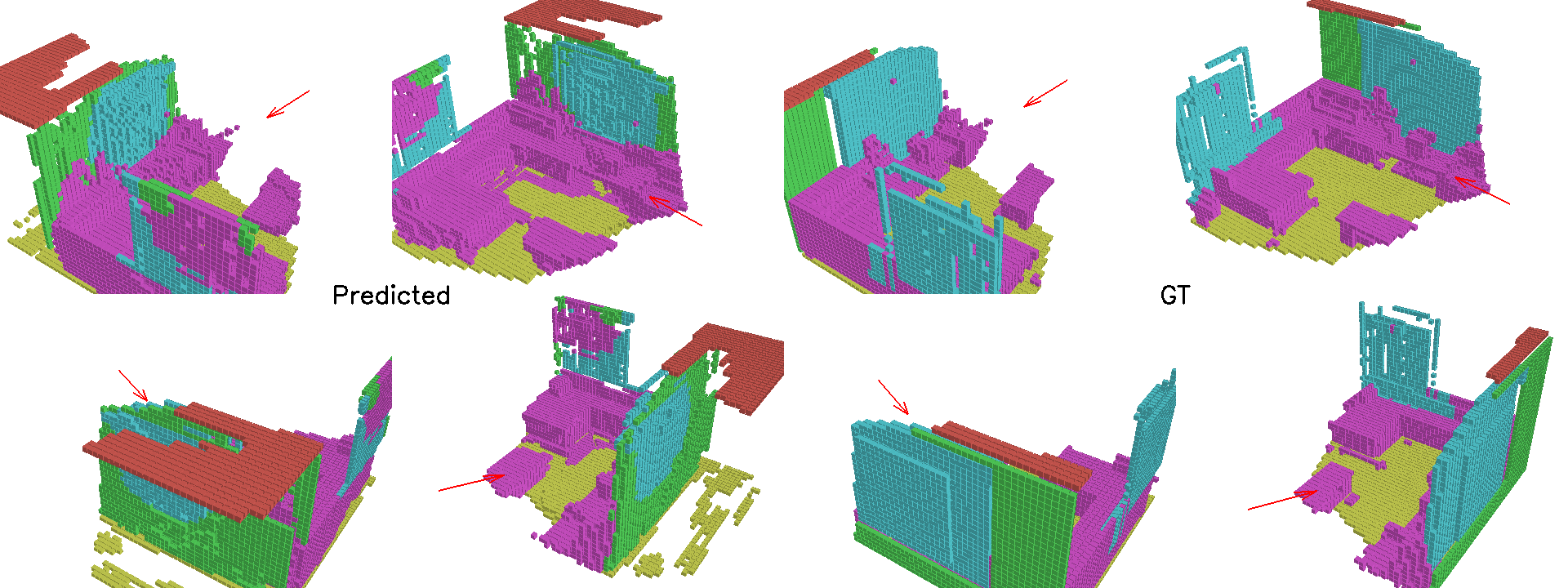}}%
\hfil
\subfloat[]{\includegraphics[width=2in, trim=0 0 1000 0, clip]{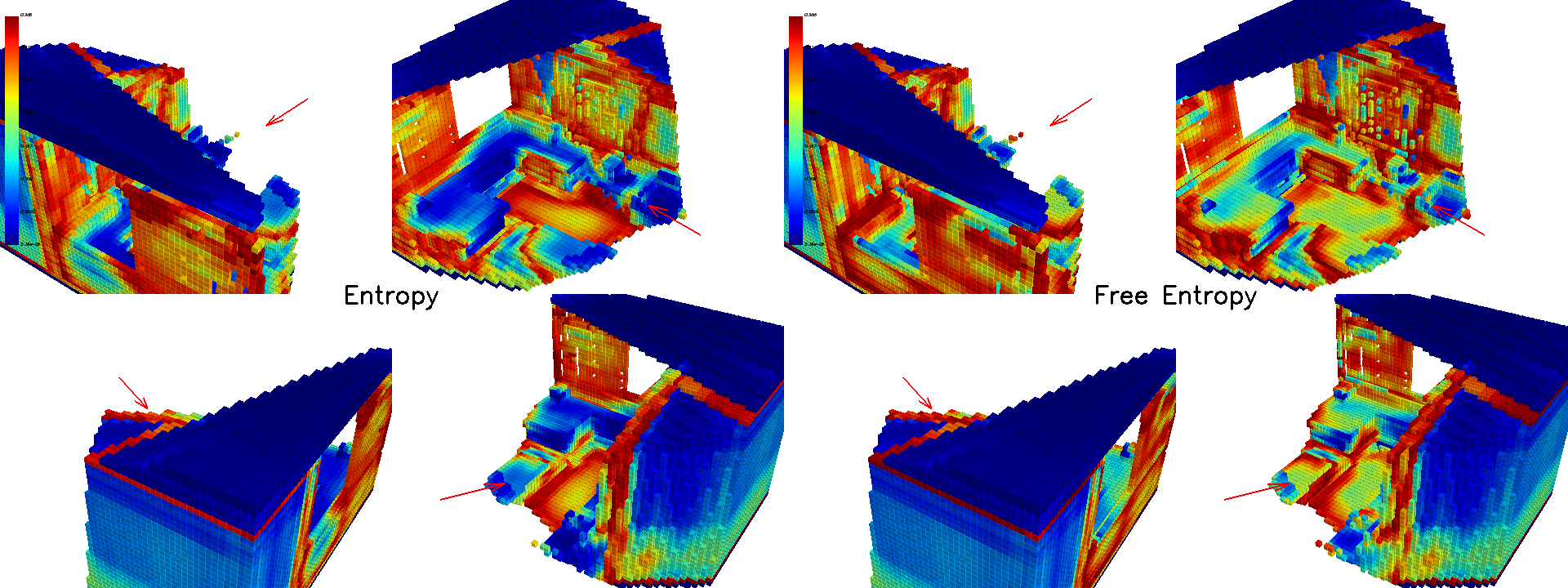}}%
\vspace{-12pt}
\hfil
\subfloat[]{\includegraphics[width=4in]{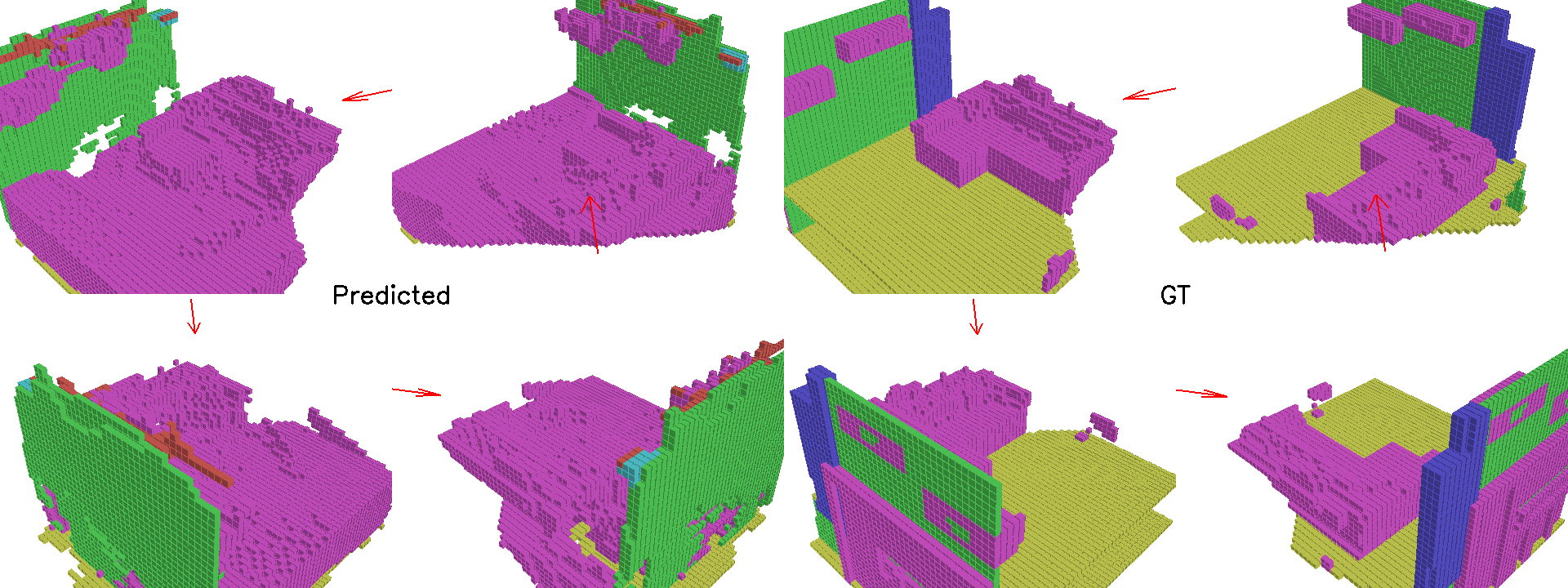}}%
\hfil
\subfloat[]{\includegraphics[width=2in, trim=0 0 1000 0, clip]{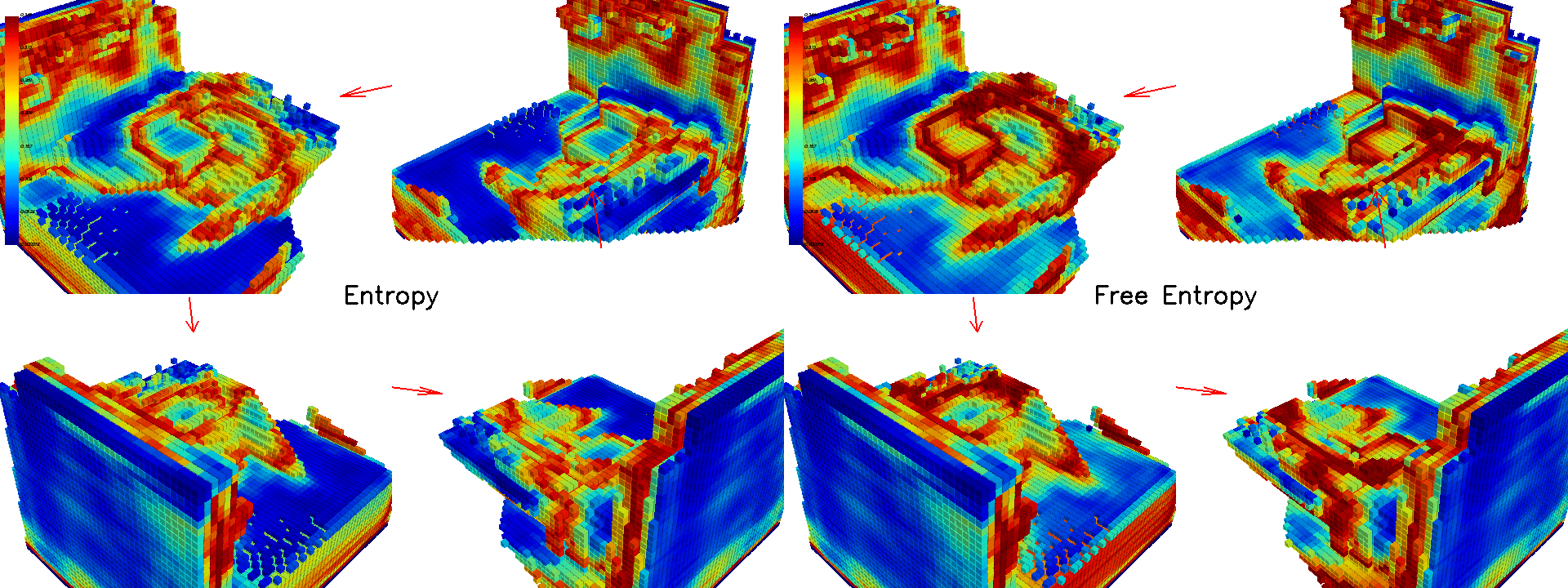}}%
\hfil
\caption{Some example output from the Bayesian SSC-Net on the SUNCG dataset. From the left we have predicted labels, ground truth and entropy.}
\label{fig:SUNCG_examples}
\end{figure*}

In Figure \ref{fig:SUNCG_examples} we show some example outputs from the Bayesian SSC-Net.

\end{document}